\documentclass[sigconf,nonacm,pdfa]{acmart}
\usepackage[a-2b]{pdfx}  

\usepackage{booktabs}
\usepackage{diagbox}
\usepackage{bm}
\usepackage{amsmath}

\usepackage{amssymb,amsfonts}
\usepackage[ruled,linesnumbered, vlined]{algorithm2e}
\usepackage[]{algpseudocode}                               
\algrenewcommand\textproc{\texttt}
\usepackage{caption}
\usepackage{bigstrut}
\usepackage{subfig}
\usepackage{enumitem}
\usepackage{float}
\usepackage{balance}
\usepackage{multirow}
\usepackage{ulem}
\usepackage{tabularx}
\usepackage{pdfpages}
\usepackage{graphicx} 
\theoremstyle{definition}

\usepackage{threeparttable}
\usepackage{makecell}

\usepackage{cleveref}
\usepackage{array}
\newcommand\vldbdoi{10.14778/3748191.3748203}
\newcommand\vldbpages{3396 - 3405}
\newcommand\vldbvolume{18}
\newcommand\vldbissue{10} 
\newcommand\vldbyear{2025}
\newcommand\vldbauthors{\authors}
\newcommand\vldbtitle{\shorttitle}
\newcommand\vldbavailabilityurl{https://github.com/TreeAI-Lab/EAGLE}
\newcommand\vldbpagestyle{empty} 
 
\begin{document}
 \title{When Speed meets Accuracy: an Efficient and Effective  Graph Model for Temporal Link Prediction}

 \author{Haoyang LI}
 \affiliation{%
 	\institution{PolyU}
 }
\email{haoyang-comp.li@polyu.edu.hk}
 
 \author{Yuming XU}
 \affiliation{%
 	\institution{PolyU \& SCUT}
 }
 \email{fm.martinx@gmail.com}
 
 \author{Yiming LI}
 \affiliation{%
 	\institution{HKUST}
 }
 \email{yliix@connect.ust.hk}

 \author{Hanmo LIU}
 \affiliation{%
 	\institution{HKUST}
 }
 \email{hliubm@connect.ust.hk}

\author{Darian LI}
\affiliation{%
	\institution{PolyU \& GDUT}
}
\email{jinlulhc@outlook.com}

\author{Chen Jason ZHANG}
\affiliation{%
	\institution{PolyU}
}
\email{jason-c.zhang@polyu.edu.hk}
 
 \author{Lei CHEN}
 \affiliation{%
 	\institution{HKUST(GZ) \& FYTRI}
 	\streetaddress{}
 }
 \email{leichen@cse.ust.hk}
 
 \author{Qing LI}
 \affiliation{%
 	\institution{PolyU}
 }
 \email{csqli@comp.polyu.edu.hk}

\begin{abstract}
Temporal link prediction in dynamic graphs is a critical task with applications in diverse domains such as social networks, recommendation systems, and e-commerce platforms. While existing Temporal Graph Neural Networks (T-GNNs) have achieved notable success by leveraging complex architectures to model temporal and structural dependencies, they often suffer from scalability and efficiency challenges due to high computational overhead.  In this paper, we propose EAGLE, a lightweight framework that integrates short-term temporal recency and long-term global structural patterns. EAGLE consists of a time-aware module that aggregates information from a node's most recent neighbors to reflect its immediate preferences, and a structure-aware module that leverages temporal personalized PageRank to capture the influence of globally important nodes. To balance these attributes, EAGLE employs an adaptive weighting mechanism to dynamically adjust their contributions based on data characteristics.  Also, EAGLE eliminates the need for complex multi-hop message passing or memory-intensive mechanisms, enabling significant improvements in  efficiency. Extensive experiments on seven real-world temporal graphs demonstrate that EAGLE consistently achieves superior performance against state-of-the-art T-GNNs in both effectiveness and efficiency, delivering more than a 50× speedup over effective transformer-based T-GNNs.
\end{abstract}

\maketitle

\pagestyle{\vldbpagestyle}
\begingroup\small\noindent\raggedright\textbf{PVLDB Reference Format:}\\
\vldbauthors. \vldbtitle. PVLDB, \vldbvolume(\vldbissue): \vldbpages, \vldbyear.\\
\href{https://doi.org/\vldbdoi}{doi:\vldbdoi}
\endgroup
\begingroup
\renewcommand\thefootnote{}\footnote{\noindent
This work is licensed under the Creative Commons BY-NC-ND 4.0 International License. Visit \url{https://creativecommons.org/licenses/by-nc-nd/4.0/} to view a copy of this license. For any use beyond those covered by this license, obtain permission by emailing \href{mailto:info@vldb.org}{info@vldb.org}. Copyright is held by the owner/author(s). Publication rights licensed to the VLDB Endowment. \\
\raggedright Proceedings of the VLDB Endowment, Vol. \vldbvolume, No. \vldbissue\ %
ISSN 2150-8097. \\
\href{https://doi.org/\vldbdoi}{doi:\vldbdoi} \\
}\addtocounter{footnote}{-1}\endgroup

\ifdefempty{\vldbavailabilityurl}{}{
\vspace{.3cm}
\begingroup\small\noindent\raggedright\textbf{PVLDB Artifact Availability:}\\
The source code, data, and/or other artifacts have been made available at \url{\vldbavailabilityurl}.
\endgroup
}

\section{Introduction}\label{sec:intro}
Temporal graphs~\cite{fan2022towards,zhang2024efficient,chen2024minimum,qin2022mining,lou2023time,DBLP:journals/pvldb/CaiKWCZLG23}, also known as dynamic graphs, represent dynamically evolving interactions, relationships, and events among nodes over time. 
These graphs are widely applied across diverse domains, including social networks~\cite{huang2016tgraph,debrouvier2021model}, recommendation systems~\cite{gao2023survey,sharma2016graphjet}, and e-commerce platforms~\cite{li2021live,li2020hierarchical}.
A critical task in temporal graph analysis is temporal link prediction~\cite{fan2022towards, wang2014link,chen2018exploiting}, 
which aims to predict whether a link (or interaction) will occur between two nodes at a specific future time based on historical data. 
Temporal link prediction is pivotal for applications,
 such as recommending friends or content in social networks, 
 predicting user-item interactions for personalized recommendations, 
 and identifying emerging patterns in dynamic systems.

Recently,   temporal graph neural networks (T-GNNs)~\cite{ pareja2020evolvegcn, huang2024retrofitting, rossi2020temporal, wang2021apan, xu2020inductive, yu2023towards,congwe2023, trivedi2019dyrep, kumar2019predicting, poursafaei2022towards},  such as DyGFormer~\cite{yu2023towards}, GraphMixer~\cite{congwe2023}, TGN~\cite{rossi2020temporal}, etc,  have achieved  success in modeling temporal graphs, 
which capture both temporal and structural information.
These T-GNNs have different model architectures, 
which leverage multiple-hop neighbor aggregation~\cite{wu2020comprehensive}, 
multi-head self-attention mechanisms~\cite{vaswani2017attention}, 
recurrent neural networks~\cite{mienye2024recurrent}, 
and memory modules~\cite{rossi2020temporal}. 
While these T-GNNs subsequently improve performance on temporal link prediction tasks, 
they often require extensive computational resources, which limits their scalability, efficiency, and generalization in real-world applications~\cite{huang2024temporal,li2023zebra}.
To address these computational challenges, researchers have explored data management techniques~\cite{li2023zebra, li2023orca, li2023early,gao2024etc, li2021cache,wang2021apan,zhou2022tgl,DBLP:journals/pvldb/ZhengWL23,yu2024genti,gao2024simple} to accelerate the   T-GNNs without sacrificing accuracy. 
For instance, methods, such as APAN~\cite{wang2021apan}, TGL~\cite{zhou2022tgl}, CacheGNN~\cite{li2021cache},  Orca~\cite{li2023orca,li2024caching}, and GENTI~\cite{yu2024genti}, 
have introduced techniques such as asynchronous message propagation, parallel sampling, 
pruning computational graphs, and caching intermediate results.

However,  despite extensive research on T-GNNs, a fundamental question remains unanswered:

\textbf{\textit{What inherent attributes of temporal graphs are truly significant for temporal link prediction, and do we really need so complex T-GNN architectures to achieve high performance?}}

In this paper, we conduct motivational experiments to highlight two orthogonal yet complementary attributes that significantly contribute to temporal link prediction: most recent neighbors and global influential nodes. Recent neighbors capture short-term temporal recency, reflecting a node's current preferences and behavioral trends. These interactions are often the strongest indicators of immediate future connections. 
For instance, in e-commerce networks like eBay or Taobao, a customer's next purchase is frequently influenced by their most recent transaction. 
Similarly, in social networks such as Twitter or Facebook, users who actively engage with specific accounts (e.g., liking or retweeting) are likely to continue interacting with them in the near future. 
In contrast, global influential nodes capture long-term dependencies and structural patterns, 
representing a node's broader context and position within the network. While these connections may not be recent, they remain impactful due to their structural significance. For example, recommendations based on highly-rated or well-connected items are often driven by global structural patterns rather than recent activity.
However, existing T-GNNs with complex architectures fail to explicitly and efficiently integrate both factors, leading to suboptimal performance and significant computational overhead.

To address these limitations, we propose EAGLE,
a lightweight and effective framework for temporal link prediction, which explicitly captures short-term temporal trends and long-term structural dependencies.
Specifically,
EAGLE consists of a time-aware module  and a structure-aware module. 
The time-aware module explicitly captures short-term trends by aggregating information from a node's top-$k_r$ most recent neighbors, 
which reflect its immediate interactions and preferences. 
Meanwhile, the structure-aware module  explicitly learns long-term dependencies by leveraging Temporal Personalized PageRank (T-PPR) to select and aggregate information from the top-$k_s$ most influential nodes in the graph. 
To balance the contributions of these modules,
EAGLE employs an adaptive weighting mechanism that dynamically adjusts the influence of temporal and structural factors based  on varying data characteristics, 
such as graphs with high temporal activity but minimal structure or vice versa.
  From the architecture perspective, EAGLE employs simple 2-layer MLPs and aggregates information from only $k_r + k_s$ nodes, enabling it to achieve significantly higher efficiency than existing T-GNNs that rely on complex architectures such as transformer and multi-hop message passing.

In summary, this work makes the following key contributions:
\begin{itemize}[leftmargin=10pt]
	\item We identify and validate two critical attributes for each node in  temporal link prediction:
	(i) most recent neighbors, which capture a node's short-term temporal recency and  preferences, and
	(ii) global influential nodes, which provide long-term global structural patterns specific to each node.
	
	\item We propose a lightweight and effective model EAGLE for temporal link prediction, 
	which explicitly integrates short-term temporal recency and long-term global structural patterns.

	\item  Extensive experiments on seven real-world temporal graph datasets show that EAGLE outperforms state-of-the-art T-GNNs in both effectiveness and efficiency.

\end{itemize}

\section{PRELIMINARY AND RELATED WORK}\label{sec:related_work}

\subsection{Temporal Graphs and Link Prediction}
Temporal graphs are very popular in real-world applications, such as Twitter and Facebook. In real-world web platforms, users may subscribe/unsubscribe to other users, join and leave the platforms, or change their attributes. These activities can be regarded as temporal graphs.
Formally, we denote a temporal graph  at time $t$ as $G(t) = (V(t), \mathbf{A}(t), \mathbf{X}(t), \mathbf{E}(t))$,
where $\mathbf{A}(t) \in \{0,1\}^{|V(t)| \times |V(t)|}$ is the adjacency matrix, $\mathbf{X}(t) \in \mathbb{R}^{|V(t)| \times d_x}$ is the node feature matrix, and $\mathbf{E}(t) \in \mathbb{R}^{n_e \times d_e}$ is the edge feature matrix.
In general, 
$G(t)$ can be constructed by a sequence of edge interaction events $\mathcal{E}(t)=\{\gamma(1), \gamma(2), \ldots, \gamma(t)\}$ arranged in ascending chronological order. 
Each interaction $\gamma(\tau_i) $ is characterized by  $\gamma(\tau_i) = (v, u, \mathbf{e}_{v,u}(\tau_i), \tau_i)$, 
where $v$ and $u$ denote nodes, $\mathbf{e}_{v,u}(\tau_i)$ is edge features, 
$\tau_i \in \mathbb{N}^+$ denotes the timestamp of $\gamma(\tau_i)$.
Particularly,
the neighbors of each $v \in V(t)$ at time $t$  
can be denoted as $N_v(t) = \{(u, \textbf{e}_{v,u}(\tau_i), \tau_i) \mid (v, u, \textbf{e}_{v,u}(\tau_i), \tau_i) \in \mathcal{E}(t), \tau _i\le t \}$.

Given a continuous-time dynamic graph $G(t)$ at  time $t$ and  two nodes $v$ and $u$,  {{the temporal link prediction}} task aims to predict the occurrence of an interaction between $v$ and $u$ after time $t$.
Temporal link prediction plays a crucial role in practical applications across diverse domains, including social networks, recommendation systems, and e-commerce. 
Early temporal link prediction methods focus on acquiring temporal embeddings with random walks (e.g., CTDNE~\cite{nguyen2018continuous}) or temporal point processes (e.g., HTNE~\cite{zuo2018embedding}, M2DNE~\cite{lu2019temporal}).
Inspired by random walks on static graphs, CTDNE~\cite{nguyen2018continuous} extends the principles of random walk-based embedding methods, 
such as DeepWalk~\cite{perozzi2014deepwalk}, node2vec~\cite{grover2016node2vec}, and LINE~\cite{tang2015line}, to learn time-dependent representations in dynamic graphs. 
However, such random walk-based approaches fail to effectively capture the joint structural and temporal dependencies in dynamic graphs, leading to the development of Temporal Graph Neural Networks~\cite{Survey-TGNN,zheng2025survey} to overcome these limitations.

\begin{table}[t]
	\centering
	
	\caption{Summary of primary notations.}
	\begin{tabular}{c|l}
		\toprule
		\textbf{Notation} & \textbf{Description} \\ \midrule
		$G$ & A continuous-time dynamic graph \\\hline
		
		$\gamma(t)$ & The interaction arriving at timestamp $t$ \\ \hline
		
		$t^-, t^+$ & The time just before and after $t$ \\ \hline
		
		$\tau,\tau_i$ & Time index \\ \hline

 		$\Delta(t)$  &   The difference between $t$ and $t^-$ \\ \hline
		
		$v,u$ & Node index \\ \hline
		
		$\mathbf{x}_{v}(t)$, $\mathbf{e}_{v,u}(t)$ & Node and edge feature\\  \hline
		
		
		$N_{v}(t)$ &  Neighbors of $v$  \\ \hline
		
		$N_{v}(t,k_r)$ &  $k_r$ most recent neighbors  from $N_v(t)$  \\ \hline

		$k_r$ &  Number of neighbors with highest recency \\ \hline
		
		$k_s$ &  Number of nodes with highest PPR scores \\ \hline
		
		
		$E_{v}(t)$ &  Edges connecting with $v$ before $t$  \\ \hline
		
		$\mathbf{h}_{v}(t)$ & Node representation\\  \hline

		
		$\pi_{v}(t)$ & T-PPR score from $v$ to all nodes \\\hline
		
		$\mathbf{P}(t)$ &Probability transition matrix \\\hline
		
		$s^{ta}_{v,u}(t)$ & Time-aware   score  \\\hline
		
		$s^{sa}_{v,u}(t)$&  Structure-aware score  \\\hline
		
		$s^{hy}_{v,u}(t)$& Hybrid score \\\hline
		
		 $d_{h}, d_{x}, d_{t}, d_e$ &   Dimensions of $\mathbf{h}_v(t)$, $\mathbf{x}_v(t)$, $\mathbf{t}$, $\mathbf{e}_{v,u}(t)$ \\

		\bottomrule
	\end{tabular}
	\label{tab:notation}
\end{table}
\setlength{\textfloatsep}{0.1cm}
\setlength{\floatsep}{0.1cm}

\subsection{Temporal Graph Neural Networks} 
\label{ssec:tgnn}
We firstly introduce the architecture design of temporal GNNs, followed by a review of existing data management approaches.

\subsubsection{Temporal GNN Architecture Design}
\label{sssec:tgnn_architecture}

To encode a node $v$ at time $t$, T-GNNs compute the representation $\mathbf{h}_v(t)$ by sampling from its neighbor set $N_v(t)$ and aggregating the features of $v$ and its neighbors, along with time encoding $\mathbf{t}$ and edge features, across $L$ layers of the network.
This process can be formalized as follows:
\begin{align}
	\label{eq:sample}
	&N_{v}(t,k) = \texttt{SAMPLE}(N_{v}(t), k), \\
	\label{eq:encode}
	&\mathbf{h}_{v}^l(t)\! =\! \texttt{ENC}\Big(\mathbf{h}_{v}^{l-1}(t^-), \mathbf{t}, \big\{\mathbf{h}_{u}^{l-1}(\tau) , \mathbf{e}_{v,u}(\tau) | (u, \tau)\! \in \! N_{v}(t)\big\}\Big),
\end{align}
where $1 \leq l \leq L$ denotes the layer index, and the initial node feature is given by $\mathbf{h}_{v}^0(t) = \mathbf{x}_v(t)$. The representation at the $L$-th layer is denoted as $\mathbf{h}_{v}(t)$.
The function $\texttt{SAMPLE}(\cdot)$ in Equation~\eqref{eq:sample} selects $k$ neighbors from $N_{v}(t)$, either randomly or based on recency. The function $\texttt{ENC}(\cdot)$ in Equation~\eqref{eq:encode} defines the specific aggregation mechanism and varies across T-GNN models. Common designs for $\texttt{ENC}(\cdot)$ include RNNs,  attention mechanisms, and transformers~\cite{vaswani2017attention}. 
Depending on the method, the inputs and configuration of $\texttt{SAMPLE}(\cdot)$ and $\texttt{ENC}(\cdot)$ may differ or be omitted \cite{kumar2019predicting,poursafaei2022towards}.

As a representative work, JODIE~\cite{kumar2019predicting} uses an RNN~\cite{RNN-Review} to update node representations based on direct edge information $\gamma(t)$. 
DyRep~\cite{trivedi2019dyrep} and TGAT~\cite{xu2020inductive} use a multi-layer self-attention mechanism to encode temporal information from neighbors.  
Recognizing the importance of capturing the dynamics within each node, memory modules are introduced in APAN~\cite{wang2021apan} and TGN~\cite{rossi2020temporal} to store and reflect node-specific historical information.  
Beyond message-passing mechanisms, CAWN~\cite{wang2022inductive} employs anonymized temporal random walks to capture temporal motifs, combined with RNNs for encoding.  
Taking a step further, Zebra~\cite{li2023zebra} adopts a time-aware personalized PageRank (T-PPR) algorithm for neighbor aggregation to learn representations.  
Moreover, TCL~\cite{wang2021tcl} and DyGFormer~\cite{yu2023towards} utilize transformers~\cite{transformer} to encode nodes, resulting in a quadratic time complexity with respect to the number of neighbors.  
To address the growing complexity of models, GraphMixer~\cite{congwe2023} uses MLPs and mean pooling to encode node and neighbor information for efficient temporal link prediction.

Unlike existing T-GNNs that use complex architectures like transformers or memory modules,  
EAGLE adopts a fundamentally different approach. EAGLE explicitly  and efficiently integrates short-term temporal recency via recent neighbors and long-term global structural patterns via influential nodes identified by T-PPR, using lightweight MLPs and a modular design. 
Therefore,   EAGLE achieves high performance while greatly reducing computational cost and memory usage.

 \subsubsection{Data Management for T-GNNs}
 \label{sssec:tgnn_data_management}
Data management techniques are orthogonal to the design of T-GNN architectures, as they accelerate T-GNNs without interfering with their core architecture~\cite{li2023early, gao2024etc, 10.14778/3632093.3632108, yu2024lsmgraph, li2021cache, wang2021apan, zhou2022tgl, DBLP:journals/pvldb/ZhengWL23, yu2024genti, gao2024simple}.
For instance, APAN~\cite{wang2021apan} propagates messages asynchronously between nodes in a non-blocking manner, efficiently updating node embeddings by reducing the overhead associated with sequential processing.
Similarly, TGL~\cite{zhou2022tgl} and GENTI~\cite{yu2024genti} employ parallel sampling techniques that enable faster neighbor sampling by leveraging distributed or parallel computation frameworks.
To address the challenges of GPU-CPU communication overhead, SIMPLE~\cite{gao2024simple} and ETC~\cite{gao2024etc} utilize data management and batch construction strategies to minimize excessive GPU-CPU data transfer time.
Additionally, caching techniques have proven effective for improving efficiency, as demonstrated by CacheGNN~\cite{li2021cache} and Orca~\cite{li2023orca, li2024caching}, which reuse frequently accessed embeddings and intermediate results in memory, thus avoiding redundant computations during neighbor aggregation and message passing.
Unlike previous data management techniques that accelerate T-GNNs using specialized sampling, caching, or parallel processing, which often bring extra complexity, EAGLE achieves efficiency through a simple and lightweight modular design.

\section{METHODS}\label{sec:method}
In this section, we begin by presenting a series of motivational experiments to highlight the critical role of time and structural information.
We introduce EAGLE, a highly efficient and effective framework for temporal graph link prediction.

\begin{figure}[t]
	\centering 	
 
	\subfloat[Wikipedia.]
	{\centering\includegraphics[width=0.48\linewidth]{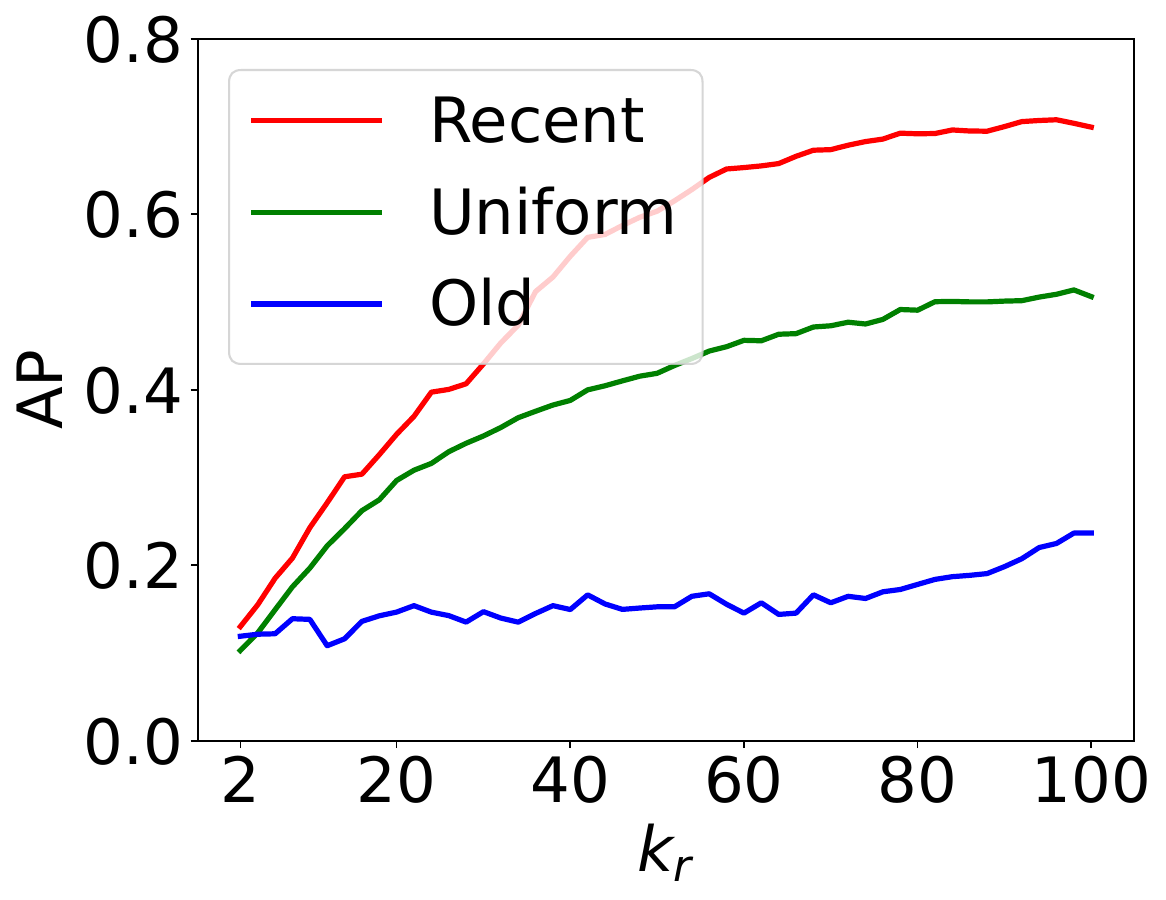}}
	\hfill
	\subfloat[AskUbuntu.]		
	{\centering\includegraphics[width=0.48\linewidth]{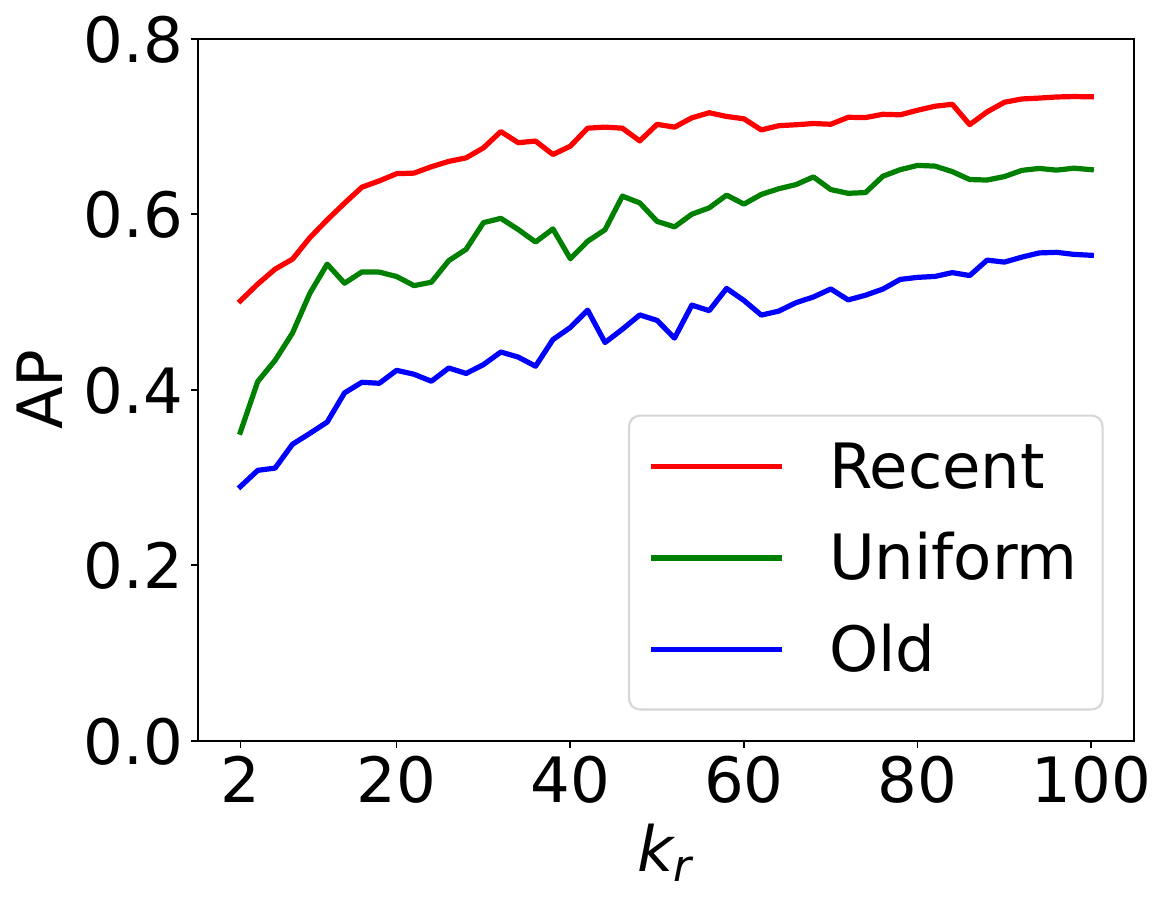}}	
 	\caption{ Time influence of neighbors.}
	\label{fig:mv1_time}
\end{figure}
\begin{figure}[t]
	\centering 	
	
	\subfloat[Cora and Citeseer graphs.]
	{\centering\includegraphics[width=0.48\linewidth]{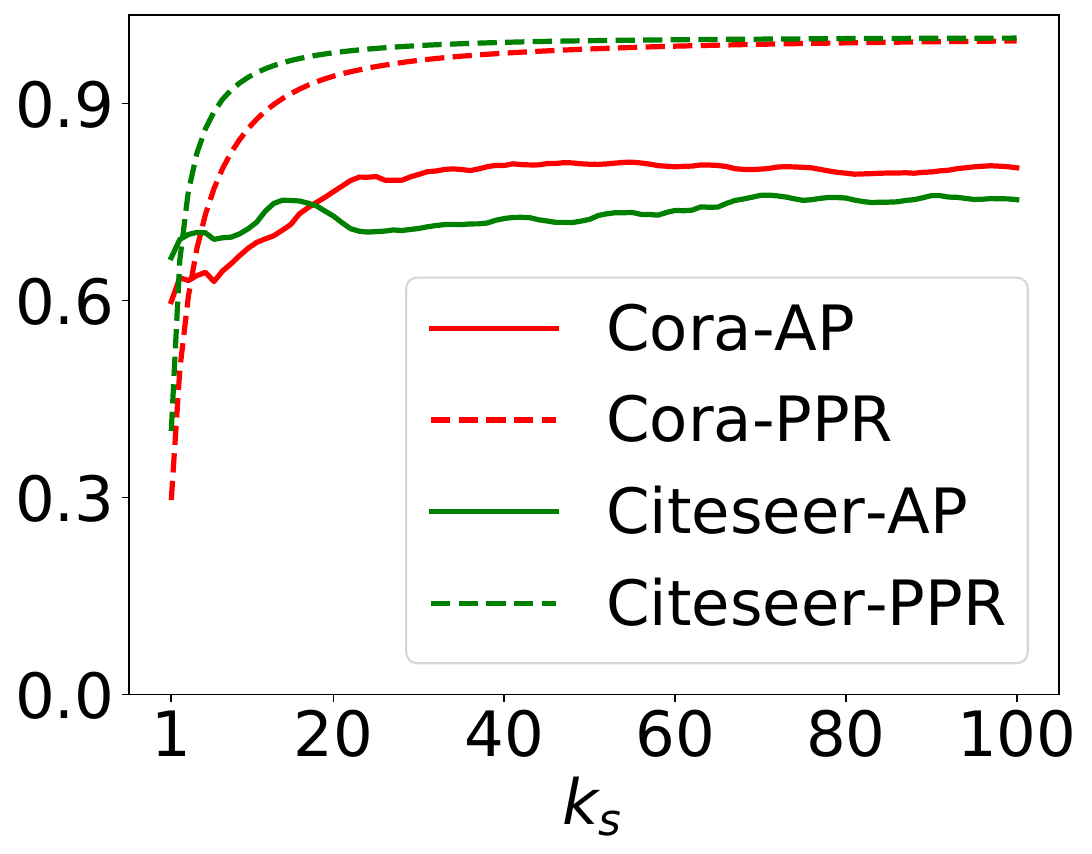}}
	\hfill
	\subfloat[Computer and Physics graphs.]		
	{\centering\includegraphics[width=0.48\linewidth]{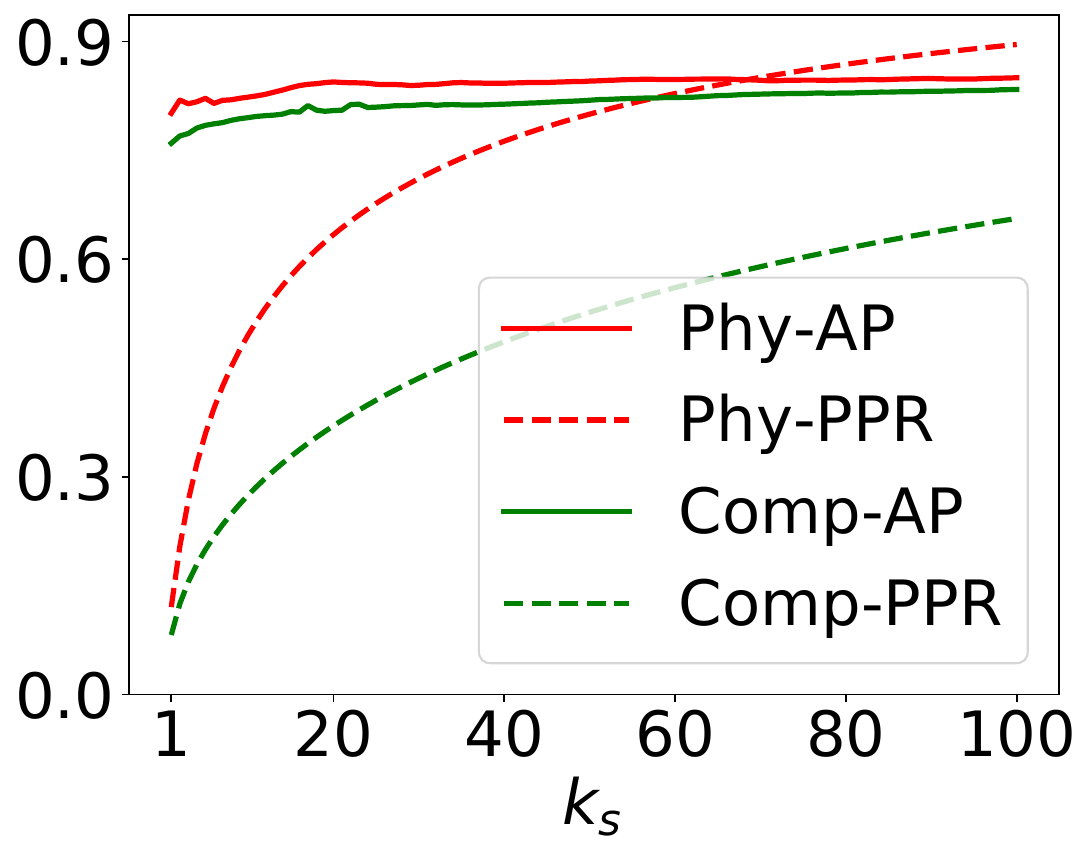}}	
	\caption{ Global influential nodes.}
	\label{fig:mv2_structure}
\end{figure}

\subsection{Motivational Observations}
\subsubsection{The Importance of Time Information}\label{sssec:me:time}
In this section, we evaluate the impact of time information on link prediction. 
Specifically, we perform link prediction experiments on two widely used temporal graphs, i.e., Wikipedia~\cite{paranjape2017motifs} and AskUbuntu~\cite{paranjape2017motifs}.
The split ratios of each graph for training, validation, and test sets are 8:1:1 in chronological order, and the evaluation metric is Average Precision (AP).
 At each time step $t$, when learning the representation for a node $v$, we use three strategies to select its $k_r$ neighbors: (1) \textbf{Recent}: select its top-$k_r$ most recent neighbors, (2) \textbf{Old}: select its top-$k_r$ oldest neighbors, and (3) \textbf{Uniform}: uniformly sample $k_r$ neighbors from all neighbors.

As shown in Figure~\ref{fig:mv1_time}~(a) and (b), 
the  {Recent} strategy consistently outperforms the  {Uniform} and  {Old} strategies on both datasets, highlighting the relevance of recent neighbors in improving model performance. 
   {Uniform}   performs moderately, while    {Old}   exhibits the lowest performance, suggesting that older interactions contribute less useful information.
 Additionally, as the number of neighbors $k_r$ increases, the performance improves across all strategies. However, the performance begins to stabilize when $k_r$ becomes sufficiently large (e.g., beyond $k_r = 40$), indicating diminishing returns from adding more neighbors. 
 These findings highlight the importance of time-aware neighbor selection and show that recent neighbors provide more valuable information for improving model effectiveness.

\subsubsection{The Importance of Structure Information}\label{sssec:me:struct}
We investigate the significance of  global graph structure for each node.
Specifically, the importance of each node $u$ to the node representation of each node $v$ is proportional to the personalized PageRank score from node $v$ to node $u$. We utilize the following theorem.
\begin{theorem}~\cite{xu2018representation}
Given a graph $G(V, \mathbf{A},\mathbf{X})$, 
a $L$-layer GCN-mean model learns node representations $\mathbf{H} \in \mathbb{R}^{|V|\times d_h}$ as follows:
\begin{align}
&	\mathbf{H}^{l+1} = \texttt{ReLU}\left( \mathbf{{A}}_n \mathbf{H}^{l} \mathbf{W}^{l} \right), \quad l = 0, 1, \dots, L-1\\
&	\mathbf{H} = \mathbf{H}^{L},\ \  \mathbf{H}^0 = \mathbf{X}
\end{align}
 where $\mathbf{A}_n = D^{-1}\mathbf{A}$ and $D$ is the degree of nodes $V$.
 Then, 
 the  importance score of node $u$ on node $v$ after $L$ propagation steps is  the expected Jacobian matrix:
$I(v, u, L) = \left\lVert \mathbb{E} \left[ \frac{\partial \mathbf{H}[v]}{\partial \mathbf{X}[u]} \right] \right\rVert_1$.
The normalized importance score $\tilde{I}(v, u, L) $ of node $u$ on $v$ is defined as:
\begin{equation}
	\tilde{I}(v, u, L) = \frac{I(v, u, L)}{\sum_{w \in \mathcal{V}} I(v, w, L)} =c\cdot \sum_{{Path}_L^{v \to u}} \prod_{i=L}^{1} \frac{1}{D[v_{i-1}]}.
\end{equation}
where $\tilde{I}(v, u, L)$ is the sum  
probabilities of all possible influential paths from $v$ to node $u$,   ${Path}_L^{v \to u}$ is a path $[v, v_{2}, \ldots, u]$ of length $L$ from node $v$ to node $u$, 
 $D[v_{i-1}]$ is the degree of $v_{i-1}$, and $c$ is an constant.
\end{theorem}

We conduct experiments on the widely used Cora~\cite{yang2016revisiting}, 
Citeseer~\cite{yang2016revisiting}, 
Amazon-Computers (COMP)~\cite{shchur2018pitfalls}, 
and Coauthor-Physics (Phy)~\cite{shchur2018pitfalls} datasets for the link prediction   with the APPNP model~\cite{gasteiger2018predict}. 
These graphs are split into training, validation, and test sets with ratios of 8:1:1. The goal is to evaluate the impact of selecting a limited number of influential neighbors for each node on the performance of the link prediction task.
To begin, we compute the Personalized PageRank (PPR) score for each pair of nodes $V$ in the graph. For each node $v \in V$, we retain only the top-$k_s$ nodes from $V$ with the highest PPR scores as its neighbors, thereby reducing the number of neighbors while focusing on the most influential ones.

As shown in Figure~\ref{fig:mv2_structure}~(a) and (b),
as $k_s$ increases, 
the average precision (AP) initially improves and then stabilizes,
even when $k_s$ is significantly smaller than the total number of nodes $|{V}|$. 
Interestingly, the average accumulated PPR scores  of the retained top-$k_s$ neighbors do not necessarily account for the largest proportion of all PPR scores. 
This observation highlights that retaining only a small number of influential nodes is sufficient.

\subsection{The EAGLE Framework}\label{ssec:egale}
In this subsection, we introduce our framework, EAGLE, for temporal link prediction, inspired by the two motivational experiments discussed earlier.
EAGLE is both highly efficient and effective at predicting links between nodes.
 It consists of two simple components, i.e., a time-aware module and a structure-aware module.

\subsubsection{Time-aware Module.}\label{sssec:time_model}
As discussed in Section~\ref{sssec:me:time}, 
from a temporal perspective, 
the most recent neighbors are more significant than older ones, as they reflect the current preferences of nodes. 
Leveraging these top-$k_r$ most recent neighbors enables the model to better capture such recent preferences.
Formally, given a graph $G(t) = (V(t), \mathbf{A}(t), \mathbf{X}(t), \mathbf{E}(t))$ and a node $v$, 
we denote the top-$k_r$ most recent neighbors of node $v$ as 
$ N_v(t, k_r) = \{(v_i, \mathbf{e}_{v,v_i}(\tau_i), \tau_i) \mid (v, v_i, \mathbf{e}_{v,v_i}(\tau_i), \tau_i),\tau_i\le t\}$, where $| N_v(t, k_r)|=k_r$.
Then, we can learn the time-aware node representation for each node $v \in V(t)$ by averaging the concatenation of the node feature $\mathbf{x}_{v_i}(t)$ and the edge feature $\mathbf{e}_{v,v_i}(\tau_i)$ of the selected nodes in $N_v(t, k_r)$ as follows:
\begin{align}\label{eq:time_aware_repre}
		\mathbf{h}_v(t) = \frac{1}{k_r}\sum_{(v_i,\mathbf{e}_{v,v_i}(\tau_i), \tau_i) \in N_v(t, k_r)} [\mathbf{x}_{v_i}(t), \mathbf{e}_{v,v_i}(\tau_i)] ,
\end{align}
After obtaining representations $\mathbf{h}_v(t)$ and $\mathbf{h}_u(t)$ for each pair of nodes $v$ and $u$ in Equation~\eqref{eq:time_aware_repre}, 
the time-aware module predicts the probability $s^{ta}_{v,u}(t) \in [0,1]$ of a link between   $v$ and $u$ as follows:
\begin{align}
	s^{ta}_{v, u}(t) = \sigma\big(\mathbf{W}_2\  \texttt{ReLU}(\mathbf{W}_1[\mathbf{h}_v(t), \mathbf{h}_u(t)]+\mathbf{b}_1) + \mathbf{b}_2\big), \label{eq:ta_score}
\end{align}
where $\sigma(\cdot)$ is  \texttt{Sigmoid} function, and $\mathbf{W}_1$, $\mathbf{b}_1$, $\mathbf{W}_2$, and $\mathbf{b}_2$ are simple trainable parameters.

\begin{figure}[t]
	\centering	
	\includegraphics[width = 0.46\textwidth]{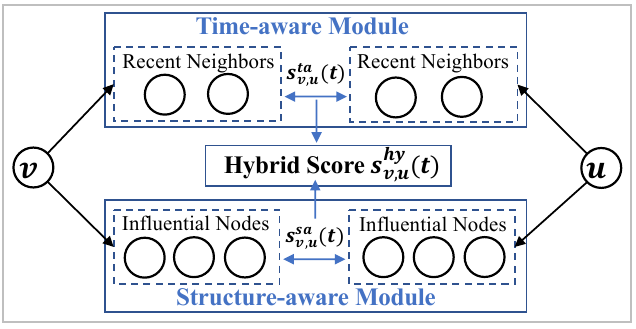}
	\caption{EAGLE framework overview. EAGLE predicts whether there is a link between nodes $v$ and $u$ at time $t$. }
	\label{fig:framework}
\end{figure}

\subsubsection{Structure-aware Module.}\label{sssec:structure_model}
As discussed in Section~\ref{sssec:me:struct}, nodes  with higher PPR scores are more important to a given node $v$.
Therefore, we employ an efficient Temporal PPR (T-PPR) algorithm~\cite{li2023zebra} to estimate the influence score of nodes in temporal graphs.
T-PPR builds upon the traditional PPR approach but extends it to dynamic graphs, where the importance of temporal information is emphasized. 
Formally,
the T-PPR~\cite{li2023zebra}  score of each  node $v$ with $\alpha$-termination probability 
 for each node $v$ is defined as follows:
\begin{align}\label{eq:ppr}
	\pi^{l+1}_{v}(t) = \alpha \cdot \mathbf{P}(t)\pi^{l}_{v}(t) + (1-\alpha) \cdot \mathbf{r}_{v}(t)
\end{align}
	where $	\pi^{l+1}_{v}(t) \in [0,1]^{|V(t)|}$ is the transition probability from  node $v$ to all nodes $V(t)$ after $l+1$ iterations
, $\pi^{0}_{v}(t)=\{1\}^{|V(t)|}$, $\alpha \in (0,1)$ is termination rate, and
$\mathbf{r}_{v}(t) \in \{0,1\}^{|V(t)|}$ is the one-hot vector of $v$, i.e., $\mathbf{r}_{v}(t)[u]=1$ if $u=v$ otherwise 0.
Also, $ \mathbf{P}(t) \in [0,1]^{|V(t)|\times |V(t)|}$ is the transition probability matrix at time $t$.
We denote the final T-PPR score of node $v$ (resp., nodes $V(t)$ ) as $\pi_{v}(t)$ (resp.,  $\pi_{V}(t)$ ) after $L$ iterations, and
$\pi_{v}(t)[u]$ denotes T-PPR score from   $v$ to   $u$ at time $t$.
Then, we introduce how to compute the   transition probability  $\mathbf{P}(t)[v_m][v_n]$.  
Given a node $v_m$ and its neighbors $N_{v_m}(t) = \{(v_n,  \mathbf{e}_{v_m,v_n}(\tau), \tau) \mid (v_m, v_n, \mathbf{e}_{v_m,v_n}(\tau), \tau) \in \mathcal{E}(t), \tau \le t \}$ at time $t$, 
the transition probability  $\mathbf{P}(t)[v_m][v_n]$ from $v_m$ to each neighbor $(v_n,\mathbf{e}_{v_m,v_n}(\tau), \tau) \in N_{v_m}(t)$ is  as:
\begin{align}\label{eq:weight}
\mathbf{P}(t)[v_m][v_n] = \frac{\beta^{|\{(v', \tau') \mid (v', \mathbf{e}_{v_m, v'}(\tau'), \tau') \in N_{v_m}(t), \tau' \geq \tau \}|}}{\sum_{z=1}^{|N_{v_m}(t)|} \beta^z},
\end{align}
where $\beta \in (0,1)$ is an exponential decay factor.
Equation~\eqref{eq:weight} ensures that more recent temporal neighbors have higher probabilities. 
Specifically, it assigns the transition probability of the $k$-th most recent temporal neighbor to be proportional to $\beta^k$.

At time $t$, given a node $v$ with its T-PPR vector $\pi_v(t)$,  nodes with the top-$k_s$ highest T-PPR scores in $\pi_v(t)$ are $N_v(\pi_v(t), k_s)=\{(v_i,\pi_v(t)[v_i])\}_{i=1}^{k_s}$.
Formally, given two nodes $v, u \in V(t)$, the set of common nodes in their top-$k_s$ T-PPR nodes
is 
$N_v(\pi_v(t), k_s)\cap N_u(\pi_u(t), k_s)$
Then, the structure-aware module predicts the probability $s^{sa}_{v,u}(t)$ of a link between node $v$ and node $u$ by summing the product of T-PPR scores of shared influential neighbors 
as follows:
\begin{align}\label{eq:sa_score}
s^{sa}_{v,u}(t) = \sum_{v_i\in N_v(\pi_v(t), k_s)\cap N_u(\pi_u(t), k_s)} \pi_v(t)[v_i] \cdot \pi_u(t)[v_i]
\end{align}
A higher   $s^{sa}_{v,u}(t)$ indicates a stronger structural connection between $v$ and $u$, increasing the likelihood of a link existing between them.

\subsubsection{Hybrid Model}\label{sssec:hybrid_model}
We discuss how to integrate the time-aware and structure-aware modules to calculate the final probability for nodes $v$ and $u$. Each module has its own limitations.
The time-aware module focuses solely on temporal recency, such as the most recent interactions of a node, but overlooks global structural patterns and long-term relevance. 
 On the other hand, the structure-aware module effectively captures global structural patterns but fails to account for time-sensitive interactions, which can result in suboptimal predictions. 
 For example, in e-commerce networks, such as eBay or Taobao,
 a customer's next purchase is often influenced by their most recent transaction.

Given each pair of nodes $v$ and $u$, the final score for determining whether an edge exists between them incorporates both the temporal score and structural score with an adaptive weighting.
Specifically, given the top-$k_r$ most recent neighbors $ N_v(t, k_r)$ of node $v$,
the average interaction time  interval between node $v$  
and its neighbors  $ N_v(t, k_r)$ can be computed  as
$\bar{t}_v = \frac{1}{k_r} \sum_{(v_i, \mathbf{e}_{v,v_i}(\tau_i), \tau_i) \in N_v(t, k_r)} (t - \tau_i).$
Smaller $\bar{t}_v$ indicates more recent interactions, while larger $\bar{t}_v$ indicates older interactions.
Therefore, given nodes $v$ and $u$, together with their time-aware score $s^{ta}_{v,u}(t)$ and structure-aware score $s^{sa}_{v,u}(t)$, 
the hybrid score $s^{hy}_{v,u}(t)$ is defined as follows:
\begin{align}\label{eq:hy_score}
s^{hy}_{v,u}(t) =\lambda \cdot \left( \exp(-{\bar{t}_v}) + \exp(-\bar{t}_u )\right) \cdot s^{ta}_{v,u}(t) + s^{sa}_{v,u}(t),
\end{align}
where 
$\lambda$ is a trade-off parameter,which can be automatically decided based on validation datasets.
The terms $\exp(-\bar{t}_v)$ and $\exp(-\bar{t}_u)$ enable the score $s^{hy}_{v,u}(t)$ to adaptively prioritize the temporal recency score $s^{ta}_{v,u}(t)$ for recent interactions (small $\bar{t}_v$ and $\bar{t}_u$) while relying more on the structural score $s^{sa}_{v,u}(t)$ for nodes with older interactions or sparse temporal activity.
Thus, the model effectively balances short-term temporal relevance and long-term structural patterns.

\subsubsection{Algorithm Summary}
Algorithm~\ref{alg:eagle}  outlines the process of predicting future links for node pairs  $\{(v_j, u_j)\}_{j=1}^m$.
Specifically, the first step of the EAGLE  is to update the graph $G(\tau_1^-)$ by incorporating the new events $\{\gamma(\tau_i)\}_{i=1}^n$, obtaining  the updated graph $G(\tau_n)$. 
Secondly, EAGLE efficiently updates the T-PPR matrix $\mathcal{\pi}_V(\tau_1^-)$ to $\mathcal{\pi}_V(\tau_n)$ based on~\cite{li2023zebra}.
Then,
for link prediction, EAGLE iterates over each node pair $(v_j, u_j)$ in $\{(v_j, u_j)\}_{j=1}^m$. 
For each pair of nodes $(v_j, u_j)$, EAGLE computes three key scores: the time-aware score $s^{ta}_{v_j, u_j}(\tau_n)$ in Equation~\eqref{eq:ta_score}, the structure-aware score $s^{sa}_{v_j, u_j}(\tau_n)$ in Equation~\eqref{eq:sa_score}, and the hybrid score $s^{hy}_{v_j, u_j}(\tau_n)$ in Equation~\eqref{eq:hy_score}.
Finally, EAGLE returns the updated graph $G(\tau_n)$, the updated T-PPR matrix $\mathcal{\pi}_V(\tau_n)$, and the hybrid  scores $\{ s^{hy}_{v_j, u_j}(\tau_n) \}_{j=1}^m$. 
 EAGLE  predicts links efficiently and effectively by leveraging a lightweight design that incorporates both temporal and structural features.
   The analysis of time and space complexity  are as follows. 
\begin{itemize}[leftmargin=10pt]
	\item \textbf{Time Complexity.}
	Given $n$ new events $\{\gamma(\tau_i)\}_{i=1}^n$, 
	EAGLE takes $O(nd_x)$ to update the graph  $G(\tau_1^-)$ to  $G(\tau_n)$ and takes
	$O(nk_s \log k_s)$ time to update the top-$k_s$ T-PPR matrix $\mathcal{\pi}_V(\tau_1^-)$ to $\mathcal{\pi}_V(\tau_n)$~\cite{li2023zebra}.
	Then, for each pair of nodes $(v_j, u_j)$, 
	EAGLE takes $O(k_rd_x+d_xd_w)$ to compute $s^{ta}_{v_j, u_j}(\tau_n)$,
	takes $O(k_s)$ to compute $s^{sa}_{v_j, u_j}(\tau_n)$, 
	and takes  $O(k_r)$ to compute $s^{hy}_{v_j, u_j}(\tau_n)$.
	Thus, 	EAGLE takes $O(n(d_x+k_s\log k_s)+m(k_rd_x+d_xd_w+k_s)$ time in total.
	\item \textbf{Space Complexity.} EAGLE only needs to store the graph $G(t)$, a single top-$k_s$ T-PPR matrix $\mathcal{\pi}_V(t)$, 
	and two MLPs in Equation~\eqref{eq:ta_score}, 
	resulting in minimal storage requirements.
	Thus, the space complexity of EAGLE is 
	$O(G) + O(k_s V(t)) + O(M)$,
	where $O(G)$ represents the storage needed for the graph $G(t)$, and $O(M)$ denotes the space required to store the model parameters.
\end{itemize}

\begin{algorithm}[t]
	\KwIn{
	Latest graph $G(\tau_1^-)$, T-PPR matrix $\mathcal{\pi}_V(\tau_1^-)$,  new events $\{\gamma(\tau_i)\}_{i=1}^n$, node pairs  $\{(v_j,u_j)\}_{j=1}^{m}$,  recent neighbor number $k_r$, and the number of top nodes ranked by T-PPR scores $k_s$.
	}
	\KwOut{ The updated graph  $G(\tau_n)$, T-PPR matrix $\mathcal{\pi}_V(\tau_n)$, and hybrid score $\{s^{hy}_{v_j,u_j}(\tau_n)\}_{j=1}^m$.
	}
	 $G(\tau_n)= G(\tau_1^-) + \{\gamma(\tau_i)\}_{i=1}^n$ \\
	 
	 $\mathcal{\pi}_V(\tau_n)  = \texttt{T-PPR\_UPDATING}(\mathcal{\pi}_V(\tau_1^-), \{\gamma(\tau_i)\}_{i=1}^n)$\\

	 \For{ $j=1$ \textbf{to} $m$}{
		 $s^{ta}_{v_j,u_j}(\tau_n) \gets \textbf{Equation}~\eqref{eq:ta_score}$ \\
		 $s^{sa}_{v_j,u_j}(\tau_n) \gets \textbf{Equation}~\eqref{eq:sa_score}$ \\
		 $s^{hy}_{v_j,u_j}(\tau_n) \gets \textbf{Equation}~\eqref{eq:hy_score}$ 
	}
	\textup{\textbf{Return\ }} $G(\tau_n)$,  $\mathcal{\pi}_V(\tau_n)$, $\{s^{hy}_{v_j,u_j}(\tau_n)\}_{j=1}^m$
	\caption{The EAGLE Framework}
	\label{alg:eagle}
\end{algorithm}
\setlength{\textfloatsep}{0.1cm}
\setlength{\floatsep}{0.1cm}

\section{EXPERIMENTS}
We compare our model EAGLE with state-of-the-art baselines on the seven graphs for link prediction.
In node classification, EAGLE outperforms all baselines in terms of effectiveness in Appendix~\ref{appx:node}

\subsection{Experimental Settings}\label{ssec:exp:setting}

\subsubsection{Datasets}\label{sssec:datasets}
Table~\ref{tab:dynamic_graph_stats} lists the seven widely-employed dynamic-graph benchmarks used in our experiments. 
Specifically, \textit{Contacts}~\cite{poursafaei2022towards} is a contact network among university students where each link weight quantifies physical proximity. 
\textit{LastFM}~\cite{poursafaei2022towards} is a bipartite user-song graph whose links denote a user playing a song. 
\textit{Wikipedia}~\cite{kumar2019predicting} and \textit{Reddit}~\cite{kumar2019predicting} track user edits and user posts, respectively, where each link represents an action. 
\textit{AskUbuntu}~\cite{paranjape2017motifs} and \textit{SuperUser}~\cite{paranjape2017motifs} are Stack Exchange traces in which a link represents a user-user exchange.  
\textit{Wiki-Talk}~\cite{paranjape2017motifs,leskovec2010governance} connects a contributor to another whenever the former edits the latter's Talk page.
Following~\cite{li2023zebra,rossi2020temporal,xu2020inductive,huang2024temporal}, 
interaction links $\mathcal{E}(t)={\gamma(1),\ldots,\gamma(t)}$ is chronologically split 70\%/15\%/15\% into training, validation, and test sets.

\subsubsection{Evaluation Metrics.} \label{sssec:metric}
	 For each test edge $(v, u, t) \in \mathcal{E}_{test}$, 
	we randomly sample $Neg_t(v)$ negative nodes to replace node $u$, treating them as negative candidates for node $v$.
	
	\noindent \underline{\textit{\textbf{Effectiveness Metric.}} }
	Following~\cite{li2023zebra,yu2023towards,congwe2023,huang2024temporal,10597888},
	we use the widely used \textbf{Average Precision (AP)},  \textbf{Mean Reciprocal Rank (MRR)},
	and
	\textbf{Hit Ratio@N (HR@N)} metrics.

	\begin{itemize}[leftmargin=*]
		\item \textbf{Average Precision (AP)}: 
		AP measures the area under  precision-recall curve, reflecting how well the model ranks true positives.

		\item \textbf{Mean Reciprocal Rank (MRR).}
		MRR is defined as
		$\mathbf{MRR} = \frac{1}{|\mathcal{E}_{\text{test}}|} \sum_{(v, u, t) \in \mathcal{E}_{\text{test}}} \frac{1}{\text{rank}(v, u, t)}$,
		where \(\text{rank}(v, u, t)\) is the rank of $u$ among the $u \cup Neg_t(v)$.

		\item \textbf{Hit Ratio@N (HR@N).}
		Given each test edge $(v,u,t)$,
		this metric evaluates whether the true positive node $u$ is ranked within the top-$N$ predictions among $u \cup Neg_t(v)$,
		which is defined as 
		$
		\textbf{HR@N} = \frac{1}{|\mathcal{E}_{\text{test}}|} \sum_{(v, u, t) \in \mathcal{E}_{\text{test}}} \mathbb{I}[\text{rank}(v, u, t) \leq N],
		$
		where \(\mathbb{I}[\text{rank}(v, u, t) \leq N]=1\) if $\text{rank}(v, u, t) \leq N$.
		
	\end{itemize}
	
	\noindent \underline{\textit{\textbf{Efficiency Metric.}}}
	We evaluate the efficiency of models  based on the total \textbf{training time (s)}, \textbf{inference time (s)},   
	\textbf{the peak GPU memory usage (MB)} in training and inference phrases.

\subsubsection{Baselines} \label{sssec:baselines}

We compare  EAGLE with six state-of-the-art T-GNNs, which are widely used~\cite{li2023zebra,huang2024temporal,yu2023towards,yi2025tgbseq,huang2024tgb2}.
 They cover  main architectures for temporal GNNs, including RNNs, transformers, and multi-hop aggregation, ensuring comprehensive comparisons.

\begin{itemize}[leftmargin=10pt]
	\item \textbf{JODIE}~\cite{kumar2019predicting} utilizes RNNs to propagate temporal interaction information, enabling  dynamic updating of node representations.
	\item \textbf{TGAT}~\cite{xu2020inductive} simulates the message-passing   of static GNNs while encoding temporal information using random Fourier features.
	\item \textbf{TGN}~\cite{rossi2020temporal} dynamically keeps a state vector for each node and generalizes previous methods\cite{kumar2019predicting, xu2020inductive} as special cases.

	\item  \textbf{GraphMixier (GMixer)}~\cite{congwe2023} encodes link and node representations via simple MLPs and mean pooling.
	
		\item \textbf{Zebra}~\cite{li2023zebra} learns node representations by aggregating neighbors with weights decided by the T-PPR process  at the target node.
		
	\item \textbf{DyGFormer}~\cite{yu2023towards} designs a transformer-based encoder to integrate time, node and edge features at once.

	\item \textbf{EAGLE (Ours)}: A lightweight framework with three variants. \textbf{EAGLE-Time} focuses on short-term temporal recency; \textbf{EAGLE-Struc.}   leverages long-term global structural patterns;   \textbf{EAGLE-Hybrid}  adaptively combines both for  temporal link prediction.

\end{itemize}

\begin{table}[t]
	\centering
 
	\caption{
		Statistics of seven dynamic graphs. $|V|$ and $|E|$ are the numbers of nodes and interactions, while $d_v$ and $d_e$ are the dimensions of node and edge features.
	}
	\label{tab:dynamic_graph_stats}
\setlength\tabcolsep{3.6pt}
\begin{tabular}{lccccc}
	\toprule
	\textbf{Dataset} & $|V|$ & $|E|$ & $d_v$ & $d_e$ & \textbf{Timespan} \\
	\midrule
	\textbf{Contacts}~\cite{poursafaei2022towards} 	& 692		& 2,426,279		& 172		& 172	& 30 days	\\
	
	\textbf{LastFM}~\cite{poursafaei2022towards} 		& 1980		& 1,293,103		& 172		& 172	& 30 days	\\
	
	\textbf{Wikipedia}~\cite{kumar2019predicting} 	& 9,227 	& 157,474 		& 172 		& 172 	& 30 days 	\\
	
	\textbf{Reddit}~\cite{kumar2019predicting} 	& 10,984 	& 672,447 		& 172 		& 172 	& 30 days 	\\
	
	\textbf{AskUbuntu}~\cite{paranjape2017motifs} 	& 159,316 	& 964,437 		& 172 		& 172 	& 2613 days \\
	
	\textbf{SuperUser}~\cite{paranjape2017motifs} 	& 194,085 	& 1,443,339 	& 172 		& 172 	& 2773 days \\
	
	\textbf{Wiki-Talk}~\cite{paranjape2017motifs,leskovec2010governance} 	& 1,140,149 & 7,833,140 	& 172 		& 172 	& 2320 days \\
	\bottomrule
\end{tabular}

\end{table}

 \subsubsection{Hyperparameter and Hardware Setting}\label{sssec:hyper_setting}
Based on validation datasets,
we tune parameters $\alpha$ in Equation~\eqref{eq:ppr} and $\beta$ in Equation~\eqref{eq:weight}  in T-PPR by $\alpha, \beta \in \{0.1, 0.2, \cdots, 1\}$,  
and tune 
 $k_r,k_s \in \{10, 20, 30, 40, 50\}$.
For the baselines,  based on validation data, we use grid search to tune the hyperparameters for each baseline on each dataset~\cite{huang2024temporal,li2023zebra}, ensuring fair comparisons.
For all models, we set the number of negative samples in the training and validation phases to 1, following~\cite{li2023zebra,rossi2020temporal,xu2020inductive,huang2024temporal}. 
For evaluation, we use 99 negative samples per node during testing, ranking the positive sample among 100 nodes. Early stopping is applied based on validation loss, with a patience of 5 epochs.
All codes are implemented by Python and  executed on a CentOS 7 machine  with a 20-core Intel Xeon Silver 4210 CPU @ 2.20GHz, 8 NVIDIA GeForce RTX 2080Ti GPUs with 22GB memory each, and 256GB of RAM.

	\begin{table*}[t]
		\vspace{-0.2em}
		\caption{Effectiveness Evaluation. \textbf{OOM} denotes out-of-GPU memory. 
			 and \textbf{TLE} indicates exceeding the two-day training limit.  
			In particular,
			EAGLE-Struc. does not have  standard deviation, as it is a training-free and deterministic algorithm in Section~\ref{sssec:structure_model}.}
		\centering
		\setlength\tabcolsep{6pt}
		\small
		
 		\begin{tabular}{c|c|c|c|c|c|c|c|ccc}
 			\hline
 			\multirow{2}{*}{\textbf{Model}} & \multirow{2}{*}{\textbf{Metric}} & \multirow{2}{*}{\textbf{JODIE}} & \multirow{2}{*}{\textbf{TGAT}} & \multirow{2}{*}{\textbf{TGN}} & \multirow{2}{*}{\textbf{GMixer}} & \multirow{2}{*}{\textbf{Zebra}} & \multirow{2}{*}{\textbf{DyGFormer}} & \multicolumn{3}{c}{\textbf{EAGLE}} \\ \cline{9-11} 
 			&                         &                         &                         &                         &                         &                         &                         & \multicolumn{1}{c|}{\textbf{Time}} & \multicolumn{1}{c|}{\textbf{Struc.}} & \textbf{Hybrid} \\ \hline

 			\multirow{3}{*}{\textbf{Contacts}}   &  \textbf{\textit{AP}}     & OOM                                & 42.30±0.63            & OOM                   & TLE                   	& 44.95±0.13                      &  TLE      & \multicolumn{1}{c|}{14.61±0.16} & \multicolumn{1}{c|}{\underline{54.20}}  	& \textbf{54.24±0.11}     \\ 
 			&  \textbf{\textit{MRR}}    & OOM                                & 68.19±0.41            & OOM                   & TLE                   										& 70.07±0.05                      & TLE      & \multicolumn{1}{c|}{60.70±0.12} & \multicolumn{1}{c|}{\underline{78.51}}  & \textbf{78.59±0.06}     \\  
 			&  \textbf{\textit{HR@10}}  & OOM                                & 96.28±0.04            & OOM                   & TLE                   										& 95.89±0.02                      & TLE      & \multicolumn{1}{c|}{78.24±0.09} & \multicolumn{1}{c|}{\underline{98.18}}  & \textbf{98.25±0.05}     \\ \hline

 			\multirow{3}{*}{\textbf{LastFM}}     &  \textbf{\textit{AP}}     & OOM                                & 13.15±0.55            & 17.73±0.48            & TLE                   	& 20.45±0.07                     & TLE      & \multicolumn{1}{c|}{20.43±0.68} & \multicolumn{1}{c|}{\underline{31.49}}  & \textbf{31.54±0.20}     \\ 
 			&  \textbf{\textit{MRR}}    & OOM                                & 23.32±0.38            & 28.05±0.32            & TLE                   										& 32.68±0.04                     & TLE      & \multicolumn{1}{c|}{31.36±0.42} & \multicolumn{1}{c|}{\underline{48.05}}  & \textbf{48.11±0.14}     \\ 
 			&  \textbf{\textit{HR@10}}  & OOM                                & 36.11±0.05            & 43.60±0.05            & TLE                   										& 47.92±0.01                     & TLE      & \multicolumn{1}{c|}{45.07±0.55} & \multicolumn{1}{c|}{\underline{67.40}}  & \textbf{67.52±0.07}     \\ \hline

 			\multirow{3}{*}{\textbf{Wikipedia}}  &  \textbf{\textit{AP}}     & 67.07±0.62                         & 66.84±1.30            & 78.96±1.21            & 69.77±0.41            	& 82.32±0.06        & \underline{86.96±0.22}            & \multicolumn{1}{c|}{69.18±0.63} & \multicolumn{1}{c|}{86.08}  & \textbf{87.02±0.29}     \\  
 			&  \textbf{\textit{MRR}}    & 71.88±0.25                         & 71.62±1.13            & 81.79±0.43            & 73.75±0.67            										& 85.56±0.04        & \underline{87.79±0.04}            & \multicolumn{1}{c|}{72.75±0.49} & \multicolumn{1}{c|}{88.09}  & \textbf{88.19±0.11}     \\  
 			&  \textbf{\textit{HR@10}}  & 85.16±0.10                         & 84.80±0.21            & 91.21±0.05            & 87.30±0.11            										& 92.10±0.02        & \textbf{92.61±0.03}            & \multicolumn{1}{c|}{84.59±0.09} & \multicolumn{1}{c|}{92.04}  & \underline{92.22±0.02}     \\ \hline 

 			\multirow{3}{*}{\textbf{Reddit}}     &  \textbf{\textit{AP}}     & 69.34±0.21                         & 72.91±0.25            & 73.37±0.91            & 70.07±0.11            	& 74.05±0.07        & \underline{77.92±0.33}            & \multicolumn{1}{c|}{31.96±0.78} & \multicolumn{1}{c|}{74.85}  & \textbf{78.09±0.66}     \\  
 			&  \textbf{\textit{MRR}}    & 75.30±0.12                         & 77.40±0.17            & 78.33±0.33            & 76.75±0.09            										& 78.89±0.05        & \underline{85.57±0.26}            & \multicolumn{1}{c|}{53.39±0.30} & \multicolumn{1}{c|}{84.23}  & \textbf{86.15±0.25}     \\  
 			&  \textbf{\textit{HR@10}}  & 91.56±0.07                         & 92.14±0.04            & 92.99±0.12            & 92.02±0.05            										& 93.23±0.01        & \underline{93.34±0.06}            & \multicolumn{1}{c|}{78.45±0.08} & \multicolumn{1}{c|}{92.53}  & \textbf{94.25±0.03}     \\ \hline
			
 			\multirow{3}{*}{\textbf{AskUbuntu}}  &  \textbf{\textit{AP}}     & OOM                                & 49.28±1.47            & 69.10±0.59            & 71.92±0.63      		& 69.02±0.17              & 69.55±0.35         & \multicolumn{1}{c|}{\underline{72.58±0.33}} & \multicolumn{1}{c|}{48.37}  & \textbf{72.80±0.19}     \\  
 			&  \textbf{\textit{MRR}}    & OOM                                & 54.83±0.69            & 73.15±0.44            & 76.91±0.38      											& 72.55±0.09              & 74.80±0.16            & \multicolumn{1}{c|}{\underline{79.44±0.25}} & \multicolumn{1}{c|}{57.58}  & \textbf{79.82±0.11}     \\ 
 			&  \textbf{\textit{HR@10}}  & OOM                                & 65.75±0.44            & 84.45±0.19            & 86.54±0.15      											& 84.13±0.02              & 85.29±0.11            & \multicolumn{1}{c|}{\underline{87.60±0.08}} & \multicolumn{1}{c|}{67.55}  & \textbf{87.72±0.02}     \\ \hline
			
 			\multirow{3}{*}{\textbf{SuperUser}}  &  \textbf{\textit{AP}}     & OOM                                & 49.95±0.89            & 67.50±0.45            & TLE                   	& 68.24±0.23                     & TLE      & \multicolumn{1}{c|}{\underline{68.63±0.91}} & \multicolumn{1}{c|}{44.89}  & \textbf{69.47±0.16}     \\ 
 			&  \textbf{\textit{MRR}}    & OOM                                & 50.92±0.38            & 72.20±0.33            & TLE                   										& 74.45±0.21                     & TLE      & \multicolumn{1}{c|}{\underline{78.09±0.77}} & \multicolumn{1}{c|}{52.96}  & \textbf{78.85±0.10}     \\  
 			&  \textbf{\textit{HR@10}}  & OOM                                & 56.57±0.10            & 81.14±0.15            & TLE                   										& 82.10±0.07                     & TLE      & \multicolumn{1}{c|}{\underline{86.02±0.29}} & \multicolumn{1}{c|}{60.69}  & \textbf{87.17±0.04}     \\ \hline

 			\multirow{3}{*}{\textbf{WikiTalk}}   &  \textbf{\textit{AP}}     & OOM                                & 58.03±0.79            & OOM                   & OOM                   	& 61.23±0.12                     & OOM      & \multicolumn{1}{c|}{\underline{66.71±0.93}} & \multicolumn{1}{c|}{46.63}  & \textbf{67.34±0.23}     \\  
 			&  \textbf{\textit{MRR}}    & OOM                                & 61.88±0.56            & OOM                   & OOM                   										& 64.43±0.09                     & OOM      & \multicolumn{1}{c|}{\underline{78.04±0.76}} & \multicolumn{1}{c|}{58.61}  & \textbf{79.15±0.17}     \\  
 			&  \textbf{\textit{HR@10}}  & OOM                                & 70.33±0.17            & OOM                   & OOM                   										& 74.78±0.05                     & OOM      & \multicolumn{1}{c|}{\underline{89.34±0.27}} & \multicolumn{1}{c|}{69.95}  & \textbf{90.28±0.05}     \\ \hline

		\end{tabular}
		\label{tab:effectiveness}
	\end{table*}

	\begin{table*}[]
		\caption{Efficiency evaluation.  
			\textbf{OOM} indicates out-of-GPU memory, and \textbf{TLE} indicates exceeding the two-day training limit.  
		}
		\centering
		\setlength\tabcolsep{6pt}
		\small
		\begin{tabular}{c|c|c|c|c|c|c|c|clc}
			\hline
			\multirow{2}{*}{\textbf{Model}} & \multirow{2}{*}{\textbf{Metric}} & \multirow{2}{*}{\textbf{JODIE}} & \multirow{2}{*}{\textbf{TGAT}} & \multirow{2}{*}{\textbf{TGN}} & \multirow{2}{*}{\textbf{GMixer}} & \multirow{2}{*}{\textbf{Zebra}} & \multirow{2}{*}{\textbf{DyGFormer}} & \multicolumn{3}{c}{\textbf{EAGLE}} \\ \cline{9-11} 
			\multicolumn{1}{c|}{}                         & \multicolumn{1}{c|}{}                        & \multicolumn{1}{c|}{}                       & \multicolumn{1}{c|}{}                      & \multicolumn{1}{c|}{}                     & \multicolumn{1}{c|}{}                        & \multicolumn{1}{c|}{}                        & \multicolumn{1}{c|}{}                       & \multicolumn{1}{c|}{\textbf{Time}}       & \multicolumn{1}{c|}{\textbf{Struc.}} & \multicolumn{1}{c}{\textbf{Hybrid}} \\ \hline
			\multirow{4}{*}{\textbf{Contacts}}                      
			&  \textbf{\textit{T-train (s)}}                                         & OOM                                         & 128457                             & OOM                                       & TLE                                          	& 9059                                          & TLE                                & \multicolumn{1}{c|}{\underline{751}}  & \multicolumn{1}{c|}{\textbf{72}}  & 823                      \\  
			&  \textbf{\textit{T-infer (s)}}                                      & OOM                                         & 40278                                   & OOM                                       & TLE                                          & 5948                                          & TLE                                     & \multicolumn{1}{c|}{\underline{240}}     & \multicolumn{1}{c|}{\textbf{210}} & 467                      \\ \cline{2-11} 
			&  \textbf{\textit{M-train (MB)}}                                       & OOM                                         & 7056                                      & OOM                                       & TLE                                     & \underline{1336}                                          & TLE                                        & \multicolumn{1}{c|}{4223}      & \multicolumn{1}{c|}{\textbf{675}}    & 4223                       \\  
			&  \textbf{\textit{M-infer (MB)}}                                    & OOM                                         & 7928                                      & OOM                                       & TLE                                        & \underline{2050}                                          & TLE                                        & \multicolumn{1}{c|}{5012}      & \multicolumn{1}{c|}{\textbf{1103}}    & 5012                       \\ \hline

			\multirow{4}{*}{\textbf{Wikipedia}}                     
			&  \textbf{\textit{T-train (s)}}                                         & 5468                                 & 4509                              & 2941                           & 10038                            									& 630                                &   6750                               & \multicolumn{1}{c|}{\underline{71}}  & \multicolumn{1}{c|}{\textbf{3}}   & 74                       \\  
			&  \textbf{\textit{T-infer (s)}}                                      & 612                                      & 541                                     & 500                                    & 1184                                      			& 411                                      &   2681                                    & \multicolumn{1}{c|}{\underline{21}}      & \multicolumn{1}{c|}{\textbf{7}}   & 29                       \\ \cline{2-11} 
			&  \textbf{\textit{M-train (MB)}}                                       & 1810                                        & 2714                                       & 3856                                      & 1440                                    & \underline{724}                                         &   1544                                       & \multicolumn{1}{c|}{1120}       & \multicolumn{1}{c|}{\textbf{340}}    & 1120                        \\  
			&  \textbf{\textit{M-infer (MB)}}                                    & 4180                                        & 4772                                       & 5226                                      & 3896                                       & \underline{2045}                                        &   3880                                      & \multicolumn{1}{c|}{3624}       & \multicolumn{1}{c|}{\textbf{560}}    & 3624                        \\ \hline

			\multirow{4}{*}{\textbf{WikiTalk}}                      
			&  \textbf{\textit{T-train (s)}}                                         & OOM                                         & 155328                              & OOM                                       & OOM                                          	& 12773                                          &  OOM                              & \multicolumn{1}{c|}{\underline{1208}} & \multicolumn{1}{c|}{\textbf{37}}  & 1245                     \\ 
			&  \textbf{\textit{T-infer (s)}}                                      & OOM                                         & 67103                                   & OOM                                       & OOM                                          & 8102                                          &  OOM                                    & \multicolumn{1}{c|}{\underline{539}}     & \multicolumn{1}{c|}{\textbf{118}} & 682                      \\ \cline{2-11} 
			&  \textbf{\textit{M-train (MB)}}                                       & OOM                                         & 14994                                      & OOM                                       & OOM                                     & \underline{2224}                                          &  OOM                                       & \multicolumn{1}{c|}{5129}       & \multicolumn{1}{c|}{\textbf{980}}    & 5129                        \\  
			&  \textbf{\textit{M-infer (MB)}}                                    & OOM                                         & 16045                                      & OOM                                       & OOM                                        & \underline{2308}                                          &  OOM                                       & \multicolumn{1}{c|}{6067}       & \multicolumn{1}{c|}{\textbf{1528}}    & 6067                        \\ \hline
		\end{tabular}
 
	\label{tab:efficiency}
	\end{table*}
	\setlength{\textfloatsep}{0.1cm}
	\setlength{\floatsep}{0.1cm}

\subsection{Main experiments}\label{ssec:exp:main_results}
\subsubsection{Effectiveness.}\label{sssec:exp:effectiveness}
We evaluate the T-GNN baselines and EAGLE on seven real-world graphs using three metrics, including AP, MRR, and HR@10. 
As shown in Table~\ref{tab:effectiveness},
our proposed EAGLE consistently achieves superior or comparable performance against all T-GNN baselines across all datasets on these three metrics, demonstrating its superior ability to balance both effectiveness and efficiency in temporal link prediction tasks.
Specifically, JODIE~\cite{kumar2019predicting}, TGAT~\cite{xu2020inductive}, 
and GraphMixer~\cite{congwe2023} achieve unsatisfactory performance, as they fail to effectively capture the critical combination of short-term temporal recency and long-term structural dependencies. 
Similarly, GraphMixer~\cite{congwe2023} uses shallow MLPs to encode node and link features among neighbors but lacks the capacity to capture global patterns effectively.
In contrast,  TGN~\cite{rossi2020temporal}, DyGFormer~\cite{yu2023towards}, and Zebra~\cite{li2023zebra} show improved performance by incorporating advanced attention mechanisms and memory modules. 
However, these models often come with significant computational overhead and fail to explicitly balance the influence of recent interactions and global structural contexts. 
For instance, Zebra~\cite{li2023zebra}  performs well in leveraging global structural dependencies using  T-PPR but overlooks the importance of temporal recency. DyGFormer~\cite{yu2023towards}, while utilizing transformers to encode node and edge features, suffers from high time and space complexity, which limits its scalability to large graphs.
Our EAGLE framework addresses these limitations by explicitly integrating short-term temporal recency and long-term global structural patterns, 
 making it robust across diverse temporal graph datasets.

\subsubsection{Efficiency.}\label{sssec:exp:efficiency}
As shown in Table~\ref{tab:efficiency},  EAGLE demonstrates significantly faster training and inference speeds while maintaining a low memory footprint, highlighting its superior scalability and practicality.
Specifically, JODIE~\cite{kumar2019predicting}, TGAT~\cite{xu2020inductive}, and GraphMixer~\cite{congwe2023} exhibit inefficiencies in both training and inference. 
JODIE relies on  inefficient RNNs,  and TGAT incorporates inefficient  temporal attention mechanisms, 
Similarly, GraphMixer requires substantial resources to process temporal and structural information, making it less efficient than EAGLE.
 DyGFormer~\cite{yu2023towards} uses transformer blocks, which are highly inefficient due to their complex self-attention mechanisms. Although Zebra~\cite{li2023zebra} improves efficiency with data management strategies, it still faces significant computational overhead due to its reliance on multiple T-PPR matrices.
EAGLE addresses these limitations with its lightweight time-aware and structure-aware modules and an efficient hybrid scoring mechanism. As a result, EAGLE delivers a speedup of over 50× compared to DyGFormer~\cite{yu2023towards}.
Peak GPU memory usage during inference is higher than during training because inference uses 99 negative samples compared to only 1 during training. Additionally, as the test progresses, the increasing graph size and expanded neighborhoods require more memory to process the growing number of nodes and neighbors.

 \vspace{-0.2em}
\subsection{Ablation Study}\label{ssec:exp:ablation}
We evaluated its \textbf{time-aware module} (EAGLE-Time), 
\textbf{structure-aware module} (EAGLE-Struc.), 
and their \textbf{hybrid model} (EAGLE-Hybrid) on both effectiveness and efficiency. 
Specifically,
regarding effectiveness, as shown in Table~\ref{tab:effectiveness},
EAGLE-Time captures short-term temporal recency by aggregating the top-$k_r$ most recent neighbors, 
excelling in datasets like AskUbuntu and SuperUser, 
where recent interactions heavily influence future behavior. 
Conversely, EAGLE-Struc., which leverages T-PPR to capture long-term structural patterns, performs well in structurally dominated datasets but fails to account for immediate temporal trends in highly dynamic graphs.
  EAGLE-Hybrid combines the strengths of both modules using an adaptive weighting mechanism, dynamically balancing short-term recency and long-term structure based on the dataset. 
This integration consistently outperforms the standalone modules in effectiveness.
In terms of efficiency, as shown in Table~\ref{tab:efficiency}, the variants of EAGLE are highly lightweight, as they process only the most recent neighbors and influential nodes, resulting in faster training and inference times. In particular, EAGLE-Struc is extremely efficient, with faster inference and lower memory usage, because it makes predictions directly from the T-PPR scores in Equation~\eqref{eq:sa_score}.

\begin{figure}[t]
	\centering 	
	\subfloat[Wikipedia.]
	{\centering\includegraphics[width=0.48\linewidth]{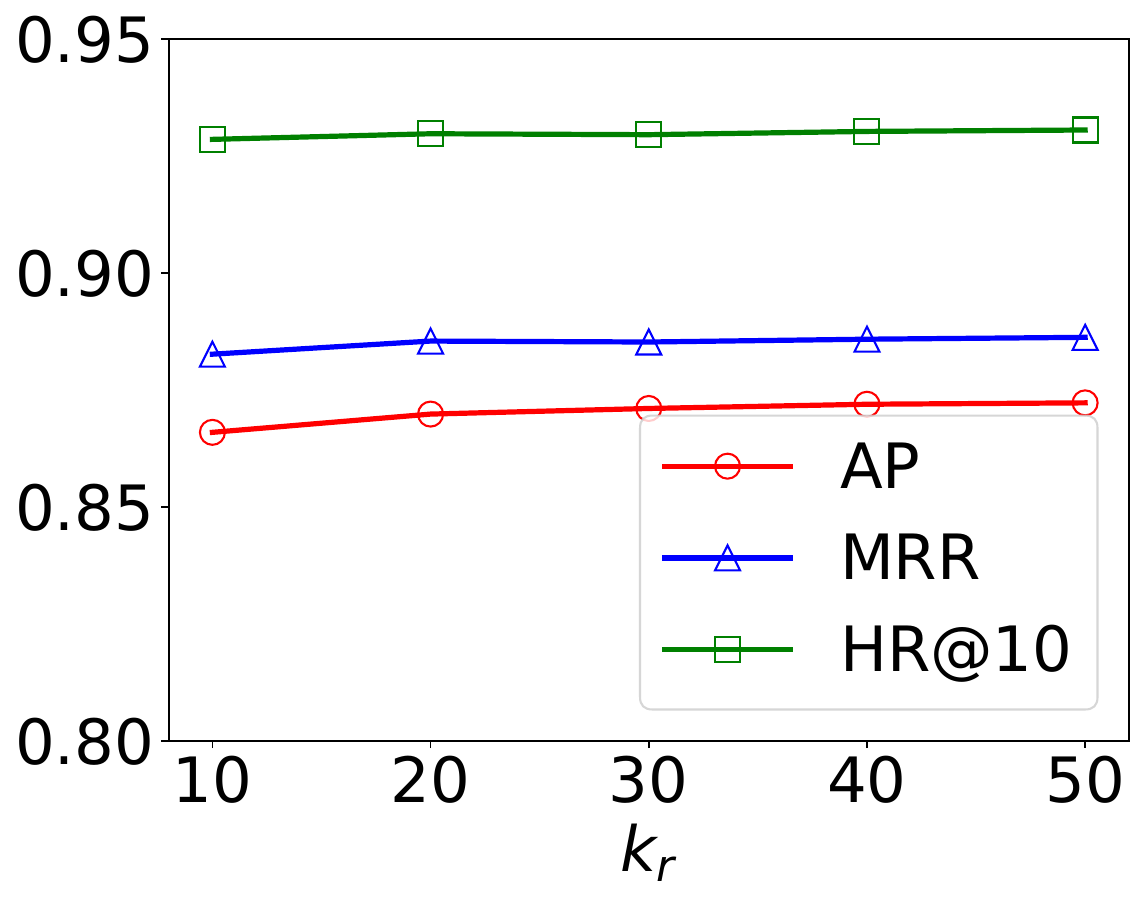}}
	\hfill
	\subfloat[AskUbuntu.]		
	{\centering\includegraphics[width=0.48\linewidth]{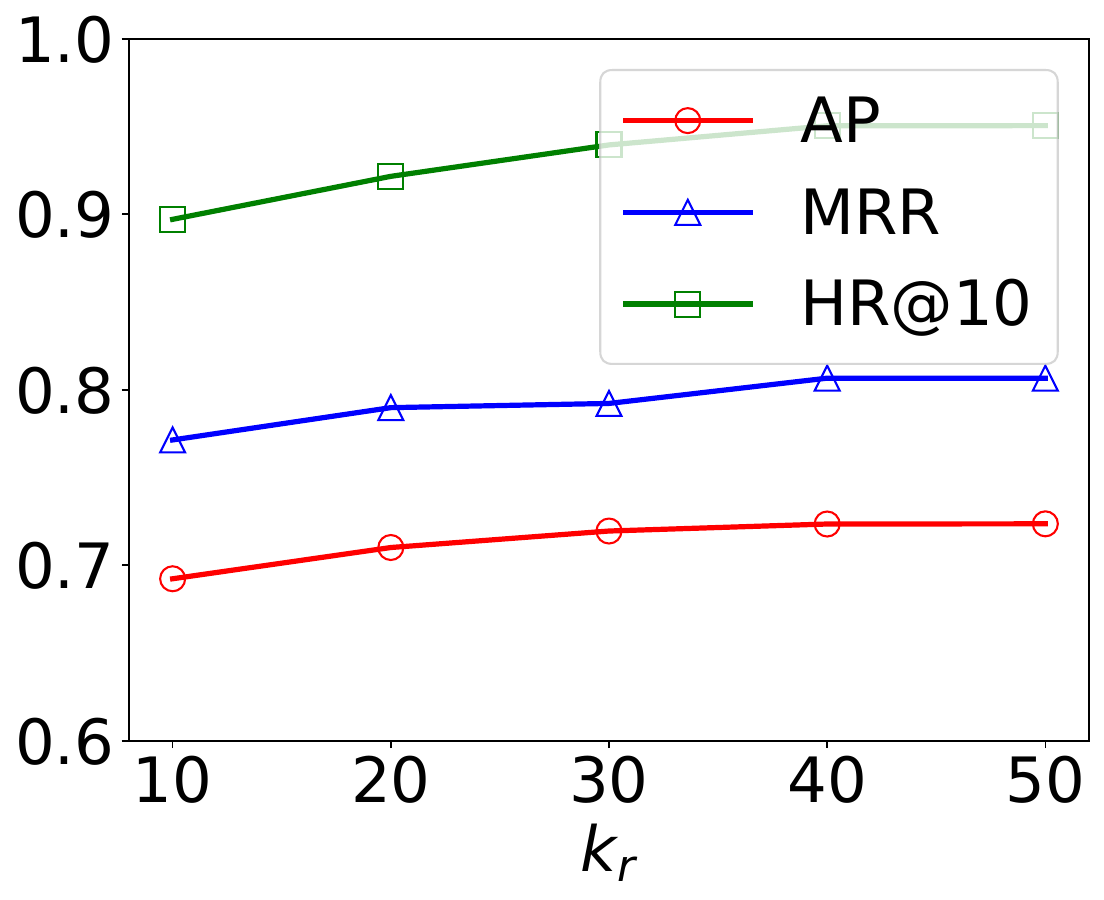}}	
	\caption{Recent neighbor number $k_r$.}
	\label{fig:param:kr}
\end{figure}

\subsection{Parameter sensitivity}\label{ssec:exp:param}
\subsubsection{Most Recent Neighbor Number $k_r$}
We analyze the impact of the number of most recent neighbors $k_r$ in Section~\ref{sssec:time_model}. 
We evaluate $k_r$ on two representative datasets, i.e., Wikipedia and AskUbuntu,  vary $k_r \in \{10,20,30,40,50\}$,  and report the metrics  AP, MRR, and HR@10.
As shown in Figure~\ref{fig:param:kr}~(a) and (b), the increasing $k_r$ generally improves model performance up to a certain threshold, beyond which the performance stabilizes. 
For instance, on the Wikipedia shown in Figure~\ref{fig:param:kr}~(a), AP, MRR, and HR@10 steadily increase as $k_r$ grows from 10 to 40, indicating that a larger number of recent neighbors provides more relevant temporal information. However, further increasing $k_r$ beyond 40 yields diminishing returns, as the added neighbors contribute little additional information.

\begin{figure}[t]
	\centering 	
	\subfloat[Wikipedia.]
	{\centering\includegraphics[width=0.48\linewidth]{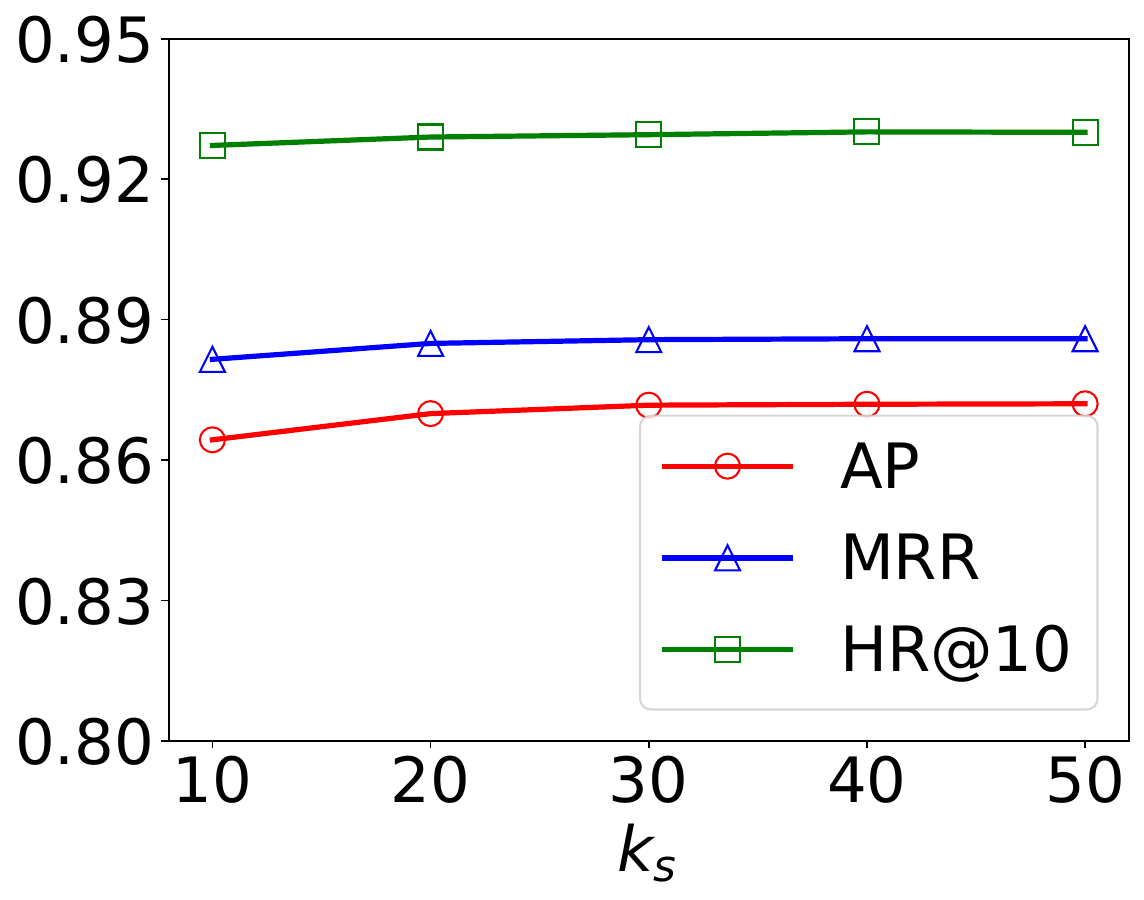}}
	\hfill
	\subfloat[Reddit.]		
	{\centering\includegraphics[width=0.48\linewidth]{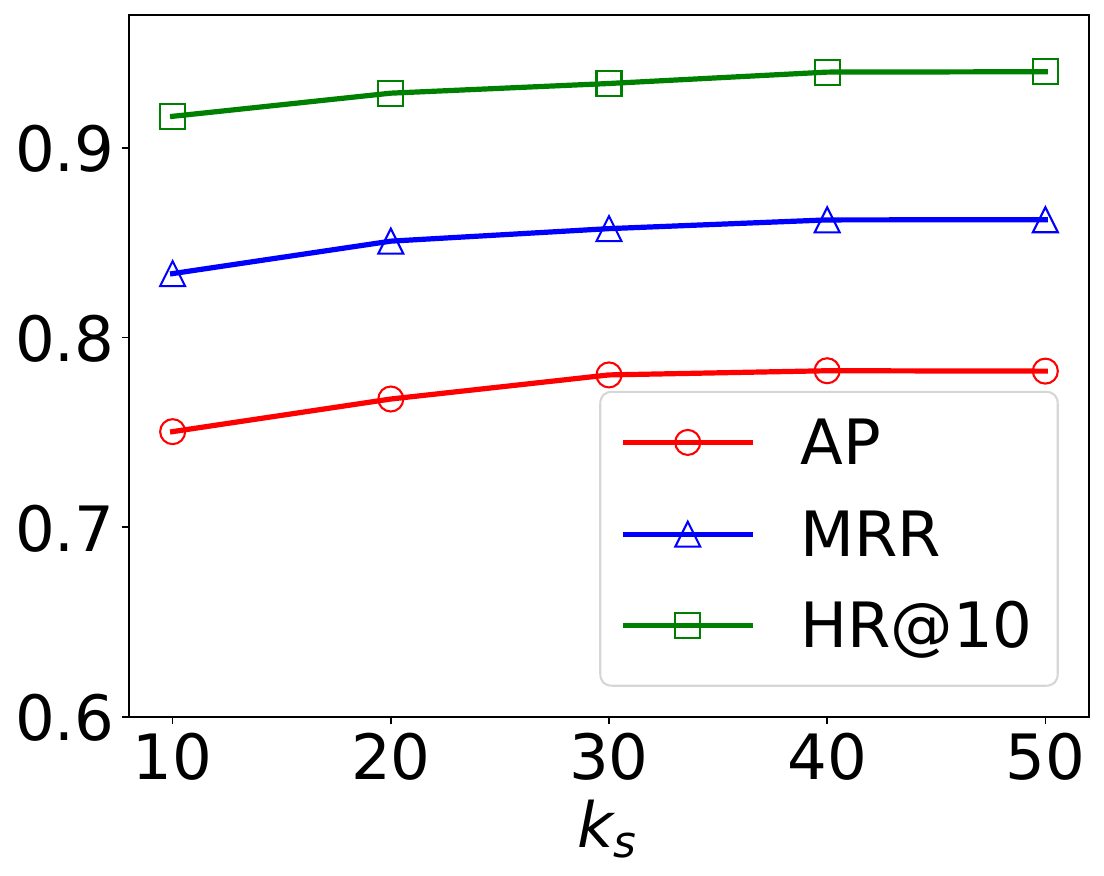}}	
	\caption{Global influential node number $k_s$.}
	\label{fig:param:ks}
\end{figure}
\subsubsection{Global Influential Node Number $k_s$}
We analyze the impact of the number of global influential nodes $k_s$  in Section~\ref{sssec:structure_model} 
for EAGLE. We evaluate the effect of $k_s$ on two representative datasets, Wikipedia and Reddit, with varying $k_s \in \{10, 20, 30, 40, 50\}$. 
The used metrics are AP, MRR, and HR@10. 
As shown in Figure~\ref{fig:param:ks}~(a) and (b),  increasing $k_s$ generally improves model performance up to a certain threshold, after which the performance stabilizes. 
For example, on the Wikipedia, AP, MRR, and HR@10 improve significantly as $k_s$ grows from 10 to 30, indicating that incorporating more globally influential nodes enriches the structural context and enhances link prediction. 
However, when $k_s$ exceeds 30, the performance gains diminish, suggesting that the most impactful structural information has already been captured.

\begin{figure}[t]
	\centering 	
	
	\subfloat[Wikipedia.]
	{\centering\includegraphics[width=0.48\linewidth]{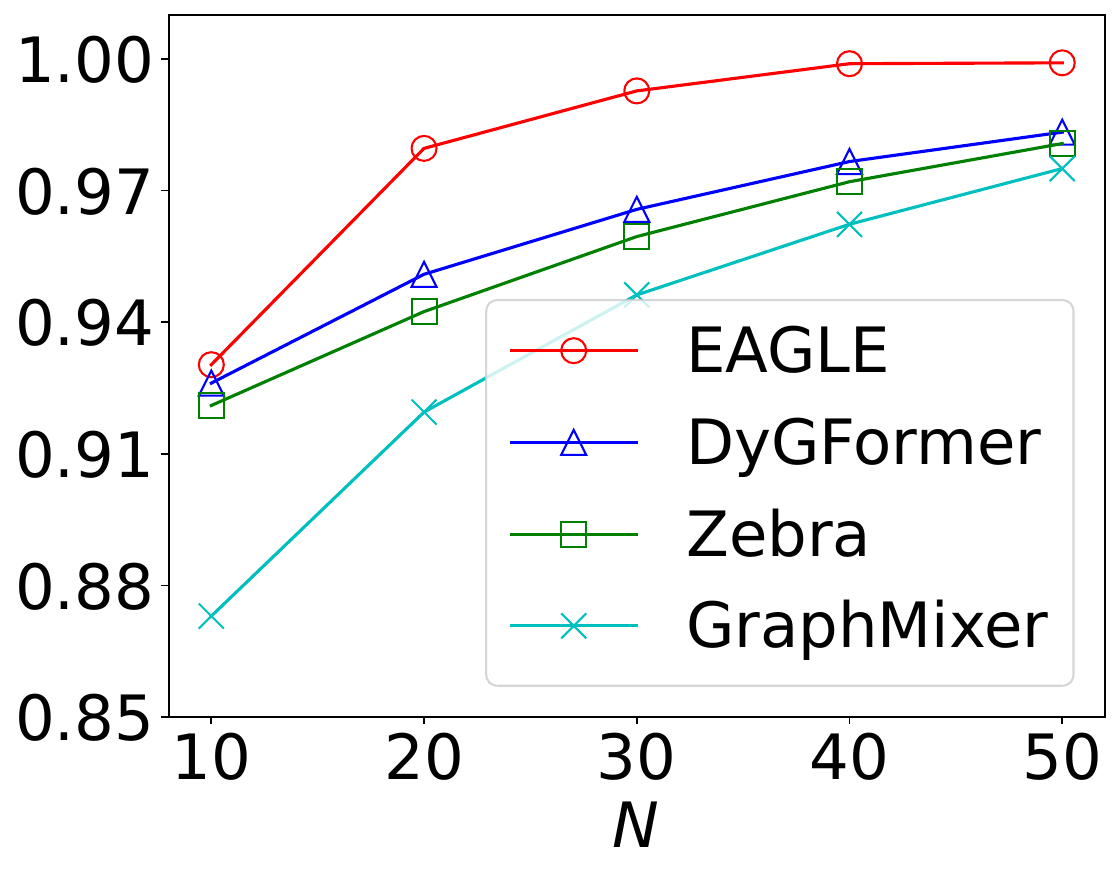}}
	\hfill
	\subfloat[AskUbuntu.]		
	{\centering\includegraphics[width=0.48\linewidth]{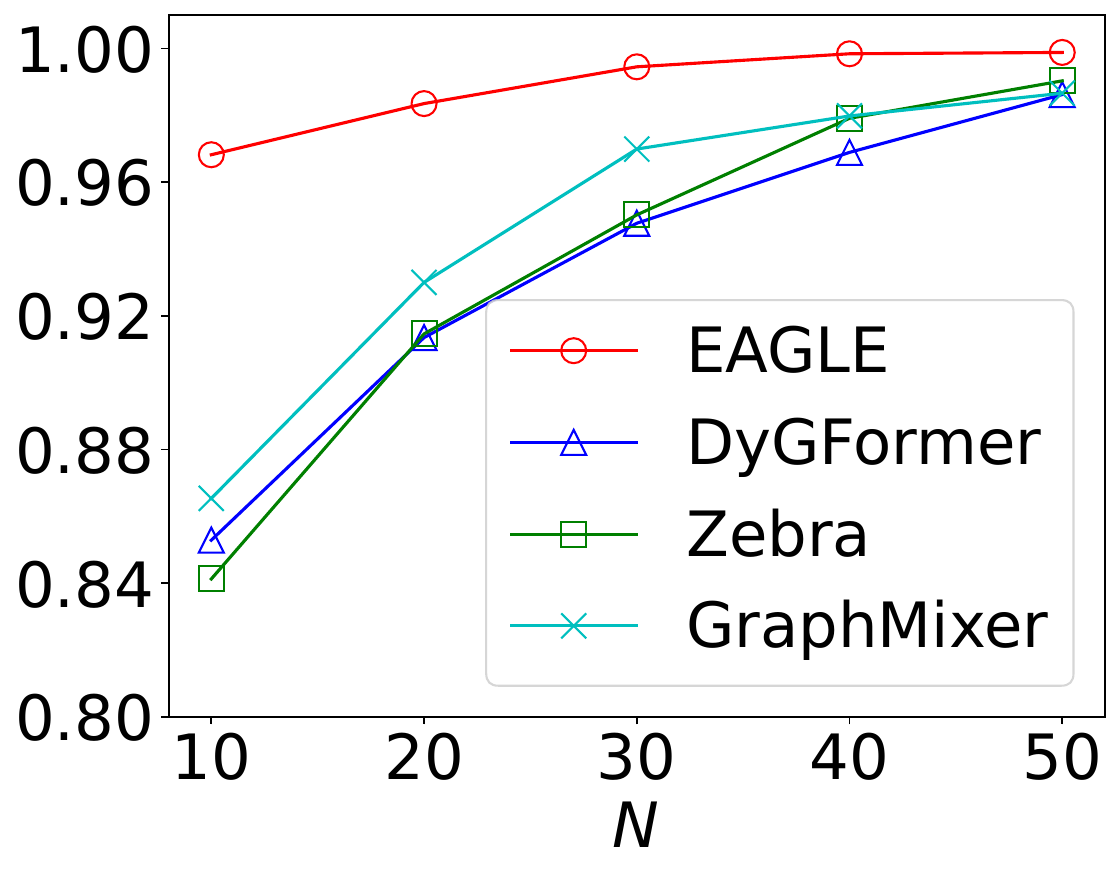}}	
	\caption{Hit Ratio@$N$.}
	\label{fig:param:hit_ratio}
\end{figure}

\subsubsection{Hit Ratio@$N$}
We compare the EAGLE model with the most effective baselines, including GMixer~\cite{congwe2023},  Zebra~\cite{li2023zebra}, and DyGFormer~\cite{yu2023towards} on HR@$N$ by varying $N \in \{10, 20, 30, 40, 50\}$. Due to similar observations, we present results on the Wikipedia and AskUbuntu datasets.
As shown in Figure~\ref{fig:param:hit_ratio}~(a) and (b), the results demonstrate that EAGLE consistently achieves higher HR@$N$ values compared to the baselines, highlighting its superior ability to prioritize true positive nodes in its predictions.
EAGLE can balance short-term temporal recency and long-term global structural patterns to deliver accurate and efficient predictions.

\section{Conclusion}\label{sec:conclusion}
We proposed  {EAGLE}, an efficient and effective framework for temporal link prediction in dynamic graphs. EAGLE integrates  {short-term temporal recency} and  {long-term global structural patterns} through a time-aware module and a structure-aware module.  
Extensive experiments on seven real-world datasets show that EAGLE outperforms various T-GNNs and is over 50 times faster than state-of-the-art transformer-based models.

 \section*{Acknowledgments}
Prof. Lei Chen's work is  supported by National Key Research and Development Program of China Grant No. 2023YFF0725100, National Science Foundation of China (NSFC) under Grant No. U22B2060, Guangdong-Hong Kong Technology Innovation Joint Funding Scheme Project No. 2024A0505040012, the Hong Kong RGC GRF Project 16213620, RIF Project R6020-19, AOE Project AoE/E-603/18, Theme-based project TRS T41-603/20R, CRF Project C2004-21G, Key Areas Special Project of Guangdong Provincial  Universities 2024ZDZX1006,  Guangdong Province Science and Technology Plan Project 2023A0505030011, Guangzhou municipality big data intelligence key lab, 2023A03J0012, Hong Kong ITC ITF grants MHX/078/21 and PRP/004/22FX, Zhujiang scholar program 2021JC02X170, Microsoft Research Asia Collaborative Research Grant, HKUST-Webank joint research lab and 2023 HKUST Shenzhen-Hong Kong Collaborative Innovation Institute Green Sustainability Special Fund, from Shui On Xintiandi and the InnoSpace GBA.
Prof. Qing Li is supported by GRF  Project 15200023 and RIF Project R1015-23.
Prof. Zhang Chen is supported by
P0045948 and P0046453 (Accel Group Holding Limited and Minshang Creative Technology Holdings Limited Donations), P0046701 and P0046703 (PolyU internal funding), P0048183 and P0048191 (Research Matching Grant funded by  UGC), P0048887 (ITF-ITSP, ITS/028/22FP), P0048984 (Postdoc Matching Fund), P0051906 (RGC Early Career Scheme, 25600624), and P0054482 (Two Square Capital Limited Donation).
Dr. Haoyang Li is supported by  research funding P0052504, P0057770, and P0053707.

\bibliographystyle{ACM-Reference-Format}

\balance

\bibliography{tgnn}


\begin{thebibliography}{66}


\ifx \showCODEN    \undefined \def \showCODEN     #1{\unskip}     \fi
\ifx \showDOI      \undefined \def \showDOI       #1{#1}\fi
\ifx \showISBNx    \undefined \def \showISBNx     #1{\unskip}     \fi
\ifx \showISBNxiii \undefined \def \showISBNxiii  #1{\unskip}     \fi
\ifx \showISSN     \undefined \def \showISSN      #1{\unskip}     \fi
\ifx \showLCCN     \undefined \def \showLCCN      #1{\unskip}     \fi
\ifx \shownote     \undefined \def \shownote      #1{#1}          \fi
\ifx \showarticletitle \undefined \def \showarticletitle #1{#1}   \fi
\ifx \showURL      \undefined \def \showURL       {\relax}        \fi
\providecommand\bibfield[2]{#2}
\providecommand\bibinfo[2]{#2}
\providecommand\natexlab[1]{#1}
\providecommand\showeprint[2][]{arXiv:#2}

\bibitem[\protect\citeauthoryear{Bertin-Mahieux, Ellis, Whitman, and
  Lamere}{Bertin-Mahieux et~al\mbox{.}}{2011}]%
        {bertin2011million}
\bibfield{author}{\bibinfo{person}{Thierry Bertin-Mahieux},
  \bibinfo{person}{Daniel~PW Ellis}, \bibinfo{person}{Brian Whitman}, {and}
  \bibinfo{person}{Paul Lamere}.} \bibinfo{year}{2011}\natexlab{}.
\newblock \showarticletitle{The million song dataset.}. In
  \bibinfo{booktitle}{\emph{Ismir}}, Vol.~\bibinfo{volume}{2}.
  \bibinfo{pages}{10}.
\newblock


\bibitem[\protect\citeauthoryear{Cai, Ke, Wang, Chen, Zhang, Liu, and Gao}{Cai
  et~al\mbox{.}}{2023}]%
        {DBLP:journals/pvldb/CaiKWCZLG23}
\bibfield{author}{\bibinfo{person}{Xin{-}Wei Cai}, \bibinfo{person}{Xiangyu
  Ke}, \bibinfo{person}{Kai Wang}, \bibinfo{person}{Lu Chen},
  \bibinfo{person}{Tianming Zhang}, \bibinfo{person}{Qing Liu}, {and}
  \bibinfo{person}{Yunjun Gao}.} \bibinfo{year}{2023}\natexlab{}.
\newblock \showarticletitle{Efficient Temporal Butterfly Counting and
  Enumeration on Temporal Bipartite Graphs}.
\newblock \bibinfo{journal}{\emph{Proc. {VLDB} Endow.}} \bibinfo{volume}{17},
  \bibinfo{number}{4} (\bibinfo{year}{2023}), \bibinfo{pages}{657--670}.
\newblock
\urldef\tempurl%
\url{https://doi.org/10.14778/3636218.3636223}
\showDOI{\tempurl}


\bibitem[\protect\citeauthoryear{Chen, Gao, Zhang, Wang, Fu, Zhang, Zhu, Gu,
  and Yu}{Chen et~al\mbox{.}}{2023}]%
        {10.14778/3632093.3632108}
\bibfield{author}{\bibinfo{person}{Chaoyi Chen}, \bibinfo{person}{Dechao Gao},
  \bibinfo{person}{Yanfeng Zhang}, \bibinfo{person}{Qiange Wang},
  \bibinfo{person}{Zhenbo Fu}, \bibinfo{person}{Xuecang Zhang},
  \bibinfo{person}{Junhua Zhu}, \bibinfo{person}{Yu Gu}, {and}
  \bibinfo{person}{Ge Yu}.} \bibinfo{year}{2023}\natexlab{}.
\newblock \showarticletitle{NeutronStream: A Dynamic GNN Training Framework
  with Sliding Window for Graph Streams}.
\newblock \bibinfo{journal}{\emph{Proc. VLDB Endow.}} \bibinfo{volume}{17},
  \bibinfo{number}{3} (\bibinfo{date}{Nov.} \bibinfo{year}{2023}),
  \bibinfo{pages}{455–468}.
\newblock
\showISSN{2150-8097}
\urldef\tempurl%
\url{https://doi.org/10.14778/3632093.3632108}
\showDOI{\tempurl}


\bibitem[\protect\citeauthoryear{Chen and Li}{Chen and Li}{2018}]%
        {chen2018exploiting}
\bibfield{author}{\bibinfo{person}{Huiyuan Chen} {and} \bibinfo{person}{Jing
  Li}.} \bibinfo{year}{2018}\natexlab{}.
\newblock \showarticletitle{Exploiting structural and temporal evolution in
  dynamic link prediction}. In \bibinfo{booktitle}{\emph{Proceedings of the
  27th ACM International conference on information and knowledge management}}.
  \bibinfo{pages}{427--436}.
\newblock


\bibitem[\protect\citeauthoryear{Chen, Shi, Peng, Lin, Wang, and Zhang}{Chen
  et~al\mbox{.}}{2024}]%
        {chen2024minimum}
\bibfield{author}{\bibinfo{person}{Xin Chen}, \bibinfo{person}{Jieming Shi},
  \bibinfo{person}{You Peng}, \bibinfo{person}{Wenqing Lin},
  \bibinfo{person}{Sibo Wang}, {and} \bibinfo{person}{Wenjie Zhang}.}
  \bibinfo{year}{2024}\natexlab{}.
\newblock \showarticletitle{Minimum Strongly Connected Subgraph Collection in
  Dynamic Graphs}.
\newblock \bibinfo{journal}{\emph{Proceedings of the VLDB Endowment}}
  \bibinfo{volume}{17}, \bibinfo{number}{6} (\bibinfo{year}{2024}),
  \bibinfo{pages}{1324--1336}.
\newblock


\bibitem[\protect\citeauthoryear{Cong, Zhang, Kang, Yuan, Wu, Zhou, Tong, and
  Mahdavi}{Cong et~al\mbox{.}}{2023}]%
        {congwe2023}
\bibfield{author}{\bibinfo{person}{Weilin Cong}, \bibinfo{person}{Si Zhang},
  \bibinfo{person}{Jian Kang}, \bibinfo{person}{Baichuan Yuan},
  \bibinfo{person}{Hao Wu}, \bibinfo{person}{Xin Zhou},
  \bibinfo{person}{Hanghang Tong}, {and} \bibinfo{person}{Mehrdad Mahdavi}.}
  \bibinfo{year}{2023}\natexlab{}.
\newblock \showarticletitle{Do We Really Need Complicated Model Architectures
  For Temporal Networks?}. In \bibinfo{booktitle}{\emph{The Eleventh
  International Conference on Learning Representations}}.
\newblock


\bibitem[\protect\citeauthoryear{Debrouvier, Parodi, Perazzo, Soliani, and
  Vaisman}{Debrouvier et~al\mbox{.}}{2021}]%
        {debrouvier2021model}
\bibfield{author}{\bibinfo{person}{Ariel Debrouvier}, \bibinfo{person}{Eliseo
  Parodi}, \bibinfo{person}{Mat{\'\i}as Perazzo}, \bibinfo{person}{Valeria
  Soliani}, {and} \bibinfo{person}{Alejandro Vaisman}.}
  \bibinfo{year}{2021}\natexlab{}.
\newblock \showarticletitle{A model and query language for temporal graph
  databases}.
\newblock \bibinfo{journal}{\emph{The VLDB Journal}} \bibinfo{volume}{30},
  \bibinfo{number}{5} (\bibinfo{year}{2021}), \bibinfo{pages}{825--858}.
\newblock


\bibitem[\protect\citeauthoryear{Fan, Jin, Lu, Tian, and Xu}{Fan
  et~al\mbox{.}}{2022}]%
        {fan2022towards}
\bibfield{author}{\bibinfo{person}{Wenfei Fan}, \bibinfo{person}{Ruochun Jin},
  \bibinfo{person}{Ping Lu}, \bibinfo{person}{Chao Tian}, {and}
  \bibinfo{person}{Ruiqi Xu}.} \bibinfo{year}{2022}\natexlab{}.
\newblock \showarticletitle{Towards event prediction in temporal graphs}.
\newblock \bibinfo{journal}{\emph{Proceedings of the VLDB Endowment}}
  \bibinfo{volume}{15}, \bibinfo{number}{9} (\bibinfo{year}{2022}),
  \bibinfo{pages}{1861--1874}.
\newblock


\bibitem[\protect\citeauthoryear{Gao, Zheng, Li, Li, Qin, Piao, Quan, Chang,
  Jin, He, et~al\mbox{.}}{Gao et~al\mbox{.}}{2023}]%
        {gao2023survey}
\bibfield{author}{\bibinfo{person}{Chen Gao}, \bibinfo{person}{Yu Zheng},
  \bibinfo{person}{Nian Li}, \bibinfo{person}{Yinfeng Li},
  \bibinfo{person}{Yingrong Qin}, \bibinfo{person}{Jinghua Piao},
  \bibinfo{person}{Yuhan Quan}, \bibinfo{person}{Jianxin Chang},
  \bibinfo{person}{Depeng Jin}, \bibinfo{person}{Xiangnan He}, {et~al\mbox{.}}}
  \bibinfo{year}{2023}\natexlab{}.
\newblock \showarticletitle{A survey of graph neural networks for recommender
  systems: Challenges, methods, and directions}.
\newblock \bibinfo{journal}{\emph{ACM Transactions on Recommender Systems}}
  \bibinfo{volume}{1}, \bibinfo{number}{1} (\bibinfo{year}{2023}),
  \bibinfo{pages}{1--51}.
\newblock


\bibitem[\protect\citeauthoryear{Gao, Li, Shen, Shao, and Chen}{Gao
  et~al\mbox{.}}{2024a}]%
        {gao2024etc}
\bibfield{author}{\bibinfo{person}{Shihong Gao}, \bibinfo{person}{Yiming Li},
  \bibinfo{person}{Yanyan Shen}, \bibinfo{person}{Yingxia Shao}, {and}
  \bibinfo{person}{Lei Chen}.} \bibinfo{year}{2024}\natexlab{a}.
\newblock \showarticletitle{ETC: Efficient Training of Temporal Graph Neural
  Networks over Large-scale Dynamic Graphs}.
\newblock \bibinfo{journal}{\emph{Proceedings of the VLDB Endowment}}
  \bibinfo{volume}{17}, \bibinfo{number}{5} (\bibinfo{year}{2024}),
  \bibinfo{pages}{1060--1072}.
\newblock


\bibitem[\protect\citeauthoryear{Gao, Li, Zhang, Shen, Shao, and Chen}{Gao
  et~al\mbox{.}}{2024b}]%
        {gao2024simple}
\bibfield{author}{\bibinfo{person}{Shihong Gao}, \bibinfo{person}{Yiming Li},
  \bibinfo{person}{Xin Zhang}, \bibinfo{person}{Yanyan Shen},
  \bibinfo{person}{Yingxia Shao}, {and} \bibinfo{person}{Lei Chen}.}
  \bibinfo{year}{2024}\natexlab{b}.
\newblock \showarticletitle{SIMPLE: Efficient Temporal Graph Neural Network
  Training at Scale with Dynamic Data Placement}.
\newblock \bibinfo{journal}{\emph{Proceedings of the ACM on Management of
  Data}} \bibinfo{volume}{2}, \bibinfo{number}{3} (\bibinfo{year}{2024}),
  \bibinfo{pages}{1--25}.
\newblock


\bibitem[\protect\citeauthoryear{Gasteiger, Bojchevski, and
  G{\"u}nnemann}{Gasteiger et~al\mbox{.}}{2018}]%
        {gasteiger2018predict}
\bibfield{author}{\bibinfo{person}{Johannes Gasteiger},
  \bibinfo{person}{Aleksandar Bojchevski}, {and} \bibinfo{person}{Stephan
  G{\"u}nnemann}.} \bibinfo{year}{2018}\natexlab{}.
\newblock \showarticletitle{Predict then propagate: Graph neural networks meet
  personalized pagerank}.
\newblock \bibinfo{journal}{\emph{arXiv preprint arXiv:1810.05997}}
  (\bibinfo{year}{2018}).
\newblock


\bibitem[\protect\citeauthoryear{Gastinger, Huang, Galkin, Loghmani, Parviz,
  Poursafaei, Danovitch, Rossi, Koutis, Stuckenschmidt, Rabbany, and
  Rabusseau}{Gastinger et~al\mbox{.}}{2024}]%
        {huang2024tgb2}
\bibfield{author}{\bibinfo{person}{Julia Gastinger}, \bibinfo{person}{Shenyang
  Huang}, \bibinfo{person}{Mikhail Galkin}, \bibinfo{person}{Erfan Loghmani},
  \bibinfo{person}{Ali Parviz}, \bibinfo{person}{Farimah Poursafaei},
  \bibinfo{person}{Jacob Danovitch}, \bibinfo{person}{Emanuele Rossi},
  \bibinfo{person}{Ioannis Koutis}, \bibinfo{person}{Heiner Stuckenschmidt},
  \bibinfo{person}{Reihaneh Rabbany}, {and} \bibinfo{person}{Guillaume
  Rabusseau}.} \bibinfo{year}{2024}\natexlab{}.
\newblock \showarticletitle{TGB 2.0: A Benchmark for Learning on Temporal
  Knowledge Graphs and Heterogeneous Graphs}.
\newblock \bibinfo{journal}{\emph{Advances in Neural Information Processing
  Systems}} (\bibinfo{year}{2024}).
\newblock


\bibitem[\protect\citeauthoryear{Grover and Leskovec}{Grover and
  Leskovec}{2016}]%
        {grover2016node2vec}
\bibfield{author}{\bibinfo{person}{Aditya Grover} {and} \bibinfo{person}{Jure
  Leskovec}.} \bibinfo{year}{2016}\natexlab{}.
\newblock \showarticletitle{node2vec: Scalable feature learning for networks}.
  In \bibinfo{booktitle}{\emph{Proceedings of the 22nd ACM SIGKDD international
  conference on Knowledge discovery and data mining}}.
  \bibinfo{pages}{855--864}.
\newblock


\bibitem[\protect\citeauthoryear{Hidasi and Tikk}{Hidasi and Tikk}{2012}]%
        {hidasi2012fast}
\bibfield{author}{\bibinfo{person}{Bal{\'a}zs Hidasi} {and}
  \bibinfo{person}{Domonkos Tikk}.} \bibinfo{year}{2012}\natexlab{}.
\newblock \showarticletitle{Fast ALS-based tensor factorization for
  context-aware recommendation from implicit feedback}. In
  \bibinfo{booktitle}{\emph{Machine Learning and Knowledge Discovery in
  Databases: European Conference, ECML PKDD 2012, Bristol, UK, September 24-28,
  2012. Proceedings, Part II 23}}. Springer, \bibinfo{pages}{67--82}.
\newblock


\bibitem[\protect\citeauthoryear{Huang, Song, Lin, Ma, and Huai}{Huang
  et~al\mbox{.}}{2016}]%
        {huang2016tgraph}
\bibfield{author}{\bibinfo{person}{Haixing Huang}, \bibinfo{person}{Jinghe
  Song}, \bibinfo{person}{Xuelian Lin}, \bibinfo{person}{Shuai Ma}, {and}
  \bibinfo{person}{Jinpeng Huai}.} \bibinfo{year}{2016}\natexlab{}.
\newblock \showarticletitle{Tgraph: A temporal graph data management system}.
  In \bibinfo{booktitle}{\emph{Proceedings of the 25th ACM International on
  Conference on Information and Knowledge Management}}.
  \bibinfo{pages}{2469--2472}.
\newblock


\bibitem[\protect\citeauthoryear{Huang, Wang, Rao, Han, Zhang, He, Xu, Zhao,
  Zheng, and Jiang}{Huang et~al\mbox{.}}{2024b}]%
        {10597888}
\bibfield{author}{\bibinfo{person}{Qiang Huang}, \bibinfo{person}{Xin Wang},
  \bibinfo{person}{Susie~Xi Rao}, \bibinfo{person}{Zhichao Han},
  \bibinfo{person}{Zitao Zhang}, \bibinfo{person}{Yongjun He},
  \bibinfo{person}{Quanqing Xu}, \bibinfo{person}{Yang Zhao},
  \bibinfo{person}{Zhigao Zheng}, {and} \bibinfo{person}{Jiawei Jiang}.}
  \bibinfo{year}{2024}\natexlab{b}.
\newblock \showarticletitle{Benchtemp: A General Benchmark for Evaluating
  Temporal Graph Neural Networks}. In \bibinfo{booktitle}{\emph{2024 IEEE 40th
  International Conference on Data Engineering (ICDE)}}.
  \bibinfo{pages}{4044--4057}.
\newblock


\bibitem[\protect\citeauthoryear{Huang, Yan, Wang, Rao, Han, Fu, Zhang, and
  Jiang}{Huang et~al\mbox{.}}{2024c}]%
        {huang2024retrofitting}
\bibfield{author}{\bibinfo{person}{Qiang Huang}, \bibinfo{person}{Xiao Yan},
  \bibinfo{person}{Xin Wang}, \bibinfo{person}{Susie~Xi Rao},
  \bibinfo{person}{Zhichao Han}, \bibinfo{person}{Fangcheng Fu},
  \bibinfo{person}{Wentao Zhang}, {and} \bibinfo{person}{Jiawei Jiang}.}
  \bibinfo{year}{2024}\natexlab{c}.
\newblock \showarticletitle{Retrofitting Temporal Graph Neural Networks with
  Transformer}.
\newblock \bibinfo{journal}{\emph{arXiv preprint arXiv:2409.05477}}
  (\bibinfo{year}{2024}).
\newblock


\bibitem[\protect\citeauthoryear{Huang, Poursafaei, Danovitch, Fey, Hu, Rossi,
  Leskovec, Bronstein, Rabusseau, and Rabbany}{Huang et~al\mbox{.}}{2024a}]%
        {huang2024temporal}
\bibfield{author}{\bibinfo{person}{Shenyang Huang}, \bibinfo{person}{Farimah
  Poursafaei}, \bibinfo{person}{Jacob Danovitch}, \bibinfo{person}{Matthias
  Fey}, \bibinfo{person}{Weihua Hu}, \bibinfo{person}{Emanuele Rossi},
  \bibinfo{person}{Jure Leskovec}, \bibinfo{person}{Michael Bronstein},
  \bibinfo{person}{Guillaume Rabusseau}, {and} \bibinfo{person}{Reihaneh
  Rabbany}.} \bibinfo{year}{2024}\natexlab{a}.
\newblock \showarticletitle{Temporal graph benchmark for machine learning on
  temporal graphs}.
\newblock \bibinfo{journal}{\emph{Advances in Neural Information Processing
  Systems}}  \bibinfo{volume}{36} (\bibinfo{year}{2024}).
\newblock


\bibitem[\protect\citeauthoryear{Kumar, Zhang, and Leskovec}{Kumar
  et~al\mbox{.}}{2019}]%
        {kumar2019predicting}
\bibfield{author}{\bibinfo{person}{Srijan Kumar}, \bibinfo{person}{Xikun
  Zhang}, {and} \bibinfo{person}{Jure Leskovec}.}
  \bibinfo{year}{2019}\natexlab{}.
\newblock \showarticletitle{Predicting dynamic embedding trajectory in temporal
  interaction networks}. In \bibinfo{booktitle}{\emph{Proceedings of the 25th
  ACM SIGKDD International Conference on Knowledge Discovery \& Data Mining}}.
  ACM, \bibinfo{pages}{1269--1278}.
\newblock


\bibitem[\protect\citeauthoryear{Leskovec, Huttenlocher, and
  Kleinberg}{Leskovec et~al\mbox{.}}{2010}]%
        {leskovec2010governance}
\bibfield{author}{\bibinfo{person}{Jure Leskovec}, \bibinfo{person}{Daniel
  Huttenlocher}, {and} \bibinfo{person}{Jon Kleinberg}.}
  \bibinfo{year}{2010}\natexlab{}.
\newblock \showarticletitle{Governance in social media: A case study of the
  Wikipedia promotion process}. In \bibinfo{booktitle}{\emph{Proceedings of the
  International AAAI Conference on Web and Social Media}},
  Vol.~\bibinfo{volume}{4}. \bibinfo{pages}{98--105}.
\newblock


\bibitem[\protect\citeauthoryear{Li and Chen}{Li and Chen}{2021}]%
        {li2021cache}
\bibfield{author}{\bibinfo{person}{Haoyang Li} {and} \bibinfo{person}{Lei
  Chen}.} \bibinfo{year}{2021}\natexlab{}.
\newblock \showarticletitle{Cache-based gnn system for dynamic graphs}. In
  \bibinfo{booktitle}{\emph{Proceedings of the 30th ACM International
  Conference on Information \& Knowledge Management}}.
  \bibinfo{pages}{937--946}.
\newblock


\bibitem[\protect\citeauthoryear{Li and Chen}{Li and Chen}{2023}]%
        {li2023early}
\bibfield{author}{\bibinfo{person}{Haoyang Li} {and} \bibinfo{person}{Lei
  Chen}.} \bibinfo{year}{2023}\natexlab{}.
\newblock \showarticletitle{Early: Efficient and reliable graph neural network
  for dynamic graphs}.
\newblock \bibinfo{journal}{\emph{Proceedings of the ACM on Management of
  Data}} \bibinfo{volume}{1}, \bibinfo{number}{2} (\bibinfo{year}{2023}),
  \bibinfo{pages}{1--28}.
\newblock


\bibitem[\protect\citeauthoryear{Li, Shen, Chen, and Yuan}{Li
  et~al\mbox{.}}{2023a}]%
        {li2023orca}
\bibfield{author}{\bibinfo{person}{Yiming Li}, \bibinfo{person}{Yanyan Shen},
  \bibinfo{person}{Lei Chen}, {and} \bibinfo{person}{Mingxuan Yuan}.}
  \bibinfo{year}{2023}\natexlab{a}.
\newblock \showarticletitle{Orca: Scalable temporal graph neural network
  training with theoretical guarantees}.
\newblock \bibinfo{journal}{\emph{Proceedings of the ACM on Management of
  Data}} \bibinfo{volume}{1}, \bibinfo{number}{1} (\bibinfo{year}{2023}),
  \bibinfo{pages}{1--27}.
\newblock


\bibitem[\protect\citeauthoryear{Li, Shen, Chen, and Yuan}{Li
  et~al\mbox{.}}{2023b}]%
        {li2023zebra}
\bibfield{author}{\bibinfo{person}{Yiming Li}, \bibinfo{person}{Yanyan Shen},
  \bibinfo{person}{Lei Chen}, {and} \bibinfo{person}{Mingxuan Yuan}.}
  \bibinfo{year}{2023}\natexlab{b}.
\newblock \showarticletitle{Zebra: When temporal graph neural networks meet
  temporal personalized PageRank}.
\newblock \bibinfo{journal}{\emph{Proceedings of the VLDB Endowment}}
  \bibinfo{volume}{16}, \bibinfo{number}{6} (\bibinfo{year}{2023}),
  \bibinfo{pages}{1332--1345}.
\newblock


\bibitem[\protect\citeauthoryear{Li, Shen, Chen, and Yuan}{Li
  et~al\mbox{.}}{2024}]%
        {li2024caching}
\bibfield{author}{\bibinfo{person}{Yiming Li}, \bibinfo{person}{Yanyan Shen},
  \bibinfo{person}{Lei Chen}, {and} \bibinfo{person}{Mingxuan Yuan}.}
  \bibinfo{year}{2024}\natexlab{}.
\newblock \showarticletitle{A Caching-based Framework for Scalable Temporal
  Graph Neural Network Training}.
\newblock \bibinfo{journal}{\emph{ACM Transactions on Database Systems}}
  (\bibinfo{year}{2024}).
\newblock


\bibitem[\protect\citeauthoryear{Li, Shen, Jiao, Pan, Zou, Meng, Yao, and
  Bu}{Li et~al\mbox{.}}{2020}]%
        {li2020hierarchical}
\bibfield{author}{\bibinfo{person}{Zhao Li}, \bibinfo{person}{Xin Shen},
  \bibinfo{person}{Yuhang Jiao}, \bibinfo{person}{Xuming Pan},
  \bibinfo{person}{Pengcheng Zou}, \bibinfo{person}{Xianling Meng},
  \bibinfo{person}{Chengwei Yao}, {and} \bibinfo{person}{Jiajun Bu}.}
  \bibinfo{year}{2020}\natexlab{}.
\newblock \showarticletitle{Hierarchical bipartite graph neural networks:
  Towards large-scale e-commerce applications}. In
  \bibinfo{booktitle}{\emph{2020 IEEE 36th International Conference on Data
  Engineering (ICDE)}}. IEEE, \bibinfo{pages}{1677--1688}.
\newblock


\bibitem[\protect\citeauthoryear{Li, Wang, Zhang, Hui, Huang, Liao, Zhang, and
  Bu}{Li et~al\mbox{.}}{2021}]%
        {li2021live}
\bibfield{author}{\bibinfo{person}{Zhao Li}, \bibinfo{person}{Haishuai Wang},
  \bibinfo{person}{Peng Zhang}, \bibinfo{person}{Pengrui Hui},
  \bibinfo{person}{Jiaming Huang}, \bibinfo{person}{Jian Liao},
  \bibinfo{person}{Ji Zhang}, {and} \bibinfo{person}{Jiajun Bu}.}
  \bibinfo{year}{2021}\natexlab{}.
\newblock \showarticletitle{Live-streaming fraud detection: A heterogeneous
  graph neural network approach}. In \bibinfo{booktitle}{\emph{Proceedings of
  the 27th ACM SIGKDD Conference on Knowledge Discovery \& Data Mining}}.
  \bibinfo{pages}{3670--3678}.
\newblock


\bibitem[\protect\citeauthoryear{Lipton, Berkowitz, and Elkan}{Lipton
  et~al\mbox{.}}{2015}]%
        {RNN-Review}
\bibfield{author}{\bibinfo{person}{Zachary~C. Lipton}, \bibinfo{person}{John
  Berkowitz}, {and} \bibinfo{person}{Charles Elkan}.}
  \bibinfo{year}{2015}\natexlab{}.
\newblock \bibinfo{title}{A Critical Review of Recurrent Neural Networks for
  Sequence Learning}.
\newblock
\newblock
\showeprint[arxiv]{1506.00019}~[cs.LG]
\urldef\tempurl%
\url{https://arxiv.org/abs/1506.00019}
\showURL{%
\tempurl}


\bibitem[\protect\citeauthoryear{Longa, Lachi, Santin, Bianchini, Lepri,
  Li{\'o}, Scarselli, and Passerini}{Longa et~al\mbox{.}}{2023}]%
        {Survey-TGNN}
\bibfield{author}{\bibinfo{person}{Antonio Longa}, \bibinfo{person}{Veronica
  Lachi}, \bibinfo{person}{Gabriele Santin}, \bibinfo{person}{Monica
  Bianchini}, \bibinfo{person}{Bruno Lepri}, \bibinfo{person}{Pietro Li{\'o}},
  \bibinfo{person}{Franco Scarselli}, {and} \bibinfo{person}{Andrea
  Passerini}.} \bibinfo{year}{2023}\natexlab{}.
\newblock \showarticletitle{Graph Neural Networks for temporal graphs: State of
  the art, open challenges, and opportunities}.
\newblock \bibinfo{journal}{\emph{ArXiv}}  \bibinfo{volume}{abs/2302.01018}
  (\bibinfo{year}{2023}).
\newblock
\urldef\tempurl%
\url{https://api.semanticscholar.org/CorpusID:256503594}
\showURL{%
\tempurl}


\bibitem[\protect\citeauthoryear{Lou, Wang, Gu, Feng, Chen, and Yu}{Lou
  et~al\mbox{.}}{2023}]%
        {lou2023time}
\bibfield{author}{\bibinfo{person}{Yunkai Lou}, \bibinfo{person}{Chaokun Wang},
  \bibinfo{person}{Tiankai Gu}, \bibinfo{person}{Hao Feng},
  \bibinfo{person}{Jun Chen}, {and} \bibinfo{person}{Jeffrey~Xu Yu}.}
  \bibinfo{year}{2023}\natexlab{}.
\newblock \showarticletitle{Time-topology analysis on temporal graphs}.
\newblock \bibinfo{journal}{\emph{The VLDB Journal}} \bibinfo{volume}{32},
  \bibinfo{number}{4} (\bibinfo{year}{2023}), \bibinfo{pages}{815--843}.
\newblock


\bibitem[\protect\citeauthoryear{Lu, Wang, Shi, Yu, and Ye}{Lu
  et~al\mbox{.}}{2019}]%
        {lu2019temporal}
\bibfield{author}{\bibinfo{person}{Yuanfu Lu}, \bibinfo{person}{Xiao Wang},
  \bibinfo{person}{Chuan Shi}, \bibinfo{person}{Philip~S Yu}, {and}
  \bibinfo{person}{Yanfang Ye}.} \bibinfo{year}{2019}\natexlab{}.
\newblock \showarticletitle{Temporal network embedding with micro-and
  macro-dynamics}. In \bibinfo{booktitle}{\emph{Proceedings of the 28th ACM
  international conference on information and knowledge management}}.
  \bibinfo{pages}{469--478}.
\newblock


\bibitem[\protect\citeauthoryear{MacDonald, Brauman, Sun, Carlson, Cassidy,
  Gerber, and West}{MacDonald et~al\mbox{.}}{2015}]%
        {macdonald2015rethinking}
\bibfield{author}{\bibinfo{person}{Graham~K MacDonald}, \bibinfo{person}{Kate~A
  Brauman}, \bibinfo{person}{Shipeng Sun}, \bibinfo{person}{Kimberly~M
  Carlson}, \bibinfo{person}{Emily~S Cassidy}, \bibinfo{person}{James~S
  Gerber}, {and} \bibinfo{person}{Paul~C West}.}
  \bibinfo{year}{2015}\natexlab{}.
\newblock \showarticletitle{Rethinking agricultural trade relationships in an
  era of globalization}.
\newblock \bibinfo{journal}{\emph{BioScience}} \bibinfo{volume}{65},
  \bibinfo{number}{3} (\bibinfo{year}{2015}), \bibinfo{pages}{275--289}.
\newblock


\bibitem[\protect\citeauthoryear{Mienye, Swart, and Obaido}{Mienye
  et~al\mbox{.}}{2024}]%
        {mienye2024recurrent}
\bibfield{author}{\bibinfo{person}{Ibomoiye~Domor Mienye},
  \bibinfo{person}{Theo~G Swart}, {and} \bibinfo{person}{George Obaido}.}
  \bibinfo{year}{2024}\natexlab{}.
\newblock \showarticletitle{Recurrent neural networks: A comprehensive review
  of architectures, variants, and applications}.
\newblock \bibinfo{journal}{\emph{Information}} \bibinfo{volume}{15},
  \bibinfo{number}{9} (\bibinfo{year}{2024}), \bibinfo{pages}{517}.
\newblock


\bibitem[\protect\citeauthoryear{Nadiri and Takes}{Nadiri and Takes}{2022}]%
        {nadiri2022large}
\bibfield{author}{\bibinfo{person}{Amirhossein Nadiri} {and}
  \bibinfo{person}{Frank~W Takes}.} \bibinfo{year}{2022}\natexlab{}.
\newblock \showarticletitle{A large-scale temporal analysis of user lifespan
  durability on the Reddit social media platform}. In
  \bibinfo{booktitle}{\emph{Companion Proceedings of the Web Conference 2022}}.
  \bibinfo{pages}{677--685}.
\newblock


\bibitem[\protect\citeauthoryear{Nguyen, Lee, Rossi, Ahmed, Koh, and
  Kim}{Nguyen et~al\mbox{.}}{2018}]%
        {nguyen2018continuous}
\bibfield{author}{\bibinfo{person}{Giang~Hoang Nguyen},
  \bibinfo{person}{John~Boaz Lee}, \bibinfo{person}{Ryan~A Rossi},
  \bibinfo{person}{Nesreen~K Ahmed}, \bibinfo{person}{Eunyee Koh}, {and}
  \bibinfo{person}{Sungchul Kim}.} \bibinfo{year}{2018}\natexlab{}.
\newblock \showarticletitle{Continuous-time dynamic network embeddings}. In
  \bibinfo{booktitle}{\emph{Companion proceedings of the the web conference
  2018}}. \bibinfo{pages}{969--976}.
\newblock


\bibitem[\protect\citeauthoryear{Paranjape, Benson, and Leskovec}{Paranjape
  et~al\mbox{.}}{2017}]%
        {paranjape2017motifs}
\bibfield{author}{\bibinfo{person}{Ashwin Paranjape}, \bibinfo{person}{Austin~R
  Benson}, {and} \bibinfo{person}{Jure Leskovec}.}
  \bibinfo{year}{2017}\natexlab{}.
\newblock \showarticletitle{Motifs in temporal networks}. In
  \bibinfo{booktitle}{\emph{Proceedings of the tenth ACM international
  conference on web search and data mining}}. \bibinfo{pages}{601--610}.
\newblock


\bibitem[\protect\citeauthoryear{Pareja, Domeniconi, Chen, Ma, Suzumura,
  Kanezashi, Kaler, Schardl, and Leiserson}{Pareja et~al\mbox{.}}{2020}]%
        {pareja2020evolvegcn}
\bibfield{author}{\bibinfo{person}{Aldo Pareja}, \bibinfo{person}{Giacomo
  Domeniconi}, \bibinfo{person}{Jie Chen}, \bibinfo{person}{Tengfei Ma},
  \bibinfo{person}{Toyotaro Suzumura}, \bibinfo{person}{Hiroki Kanezashi},
  \bibinfo{person}{Tim Kaler}, \bibinfo{person}{Tao Schardl}, {and}
  \bibinfo{person}{Charles Leiserson}.} \bibinfo{year}{2020}\natexlab{}.
\newblock \showarticletitle{Evolvegcn: Evolving graph convolutional networks
  for dynamic graphs}. In \bibinfo{booktitle}{\emph{Proceedings of the AAAI
  conference on artificial intelligence}}, Vol.~\bibinfo{volume}{34}.
  \bibinfo{pages}{5363--5370}.
\newblock


\bibitem[\protect\citeauthoryear{Perozzi, Al-Rfou, and Skiena}{Perozzi
  et~al\mbox{.}}{2014}]%
        {perozzi2014deepwalk}
\bibfield{author}{\bibinfo{person}{Bryan Perozzi}, \bibinfo{person}{Rami
  Al-Rfou}, {and} \bibinfo{person}{Steven Skiena}.}
  \bibinfo{year}{2014}\natexlab{}.
\newblock \showarticletitle{Deepwalk: Online learning of social
  representations}. In \bibinfo{booktitle}{\emph{Proceedings of the 20th ACM
  SIGKDD international conference on Knowledge discovery and data mining}}.
  \bibinfo{pages}{701--710}.
\newblock


\bibitem[\protect\citeauthoryear{Poursafaei, Huang, Pelrine, and
  Rabbany}{Poursafaei et~al\mbox{.}}{2022}]%
        {poursafaei2022towards}
\bibfield{author}{\bibinfo{person}{Farimah Poursafaei},
  \bibinfo{person}{Shenyang Huang}, \bibinfo{person}{Kellin Pelrine}, {and}
  \bibinfo{person}{Reihaneh Rabbany}.} \bibinfo{year}{2022}\natexlab{}.
\newblock \showarticletitle{Towards better evaluation for dynamic link
  prediction}.
\newblock \bibinfo{journal}{\emph{Advances in Neural Information Processing
  Systems}}  \bibinfo{volume}{35} (\bibinfo{year}{2022}),
  \bibinfo{pages}{32928--32941}.
\newblock


\bibitem[\protect\citeauthoryear{Qin, Li, Yuan, Wang, Qin, and Zhang}{Qin
  et~al\mbox{.}}{2022}]%
        {qin2022mining}
\bibfield{author}{\bibinfo{person}{Hongchao Qin}, \bibinfo{person}{Rong-Hua
  Li}, \bibinfo{person}{Ye Yuan}, \bibinfo{person}{Guoren Wang},
  \bibinfo{person}{Lu Qin}, {and} \bibinfo{person}{Zhiwei Zhang}.}
  \bibinfo{year}{2022}\natexlab{}.
\newblock \showarticletitle{Mining bursting core in large temporal graphs}.
\newblock \bibinfo{journal}{\emph{Proceedings of the VLDB Endowment}}
  (\bibinfo{year}{2022}).
\newblock


\bibitem[\protect\citeauthoryear{Rossi, Chamberlain, Frasca, Eynard, Monti, and
  Bronstein}{Rossi et~al\mbox{.}}{2020}]%
        {rossi2020temporal}
\bibfield{author}{\bibinfo{person}{Emanuele Rossi}, \bibinfo{person}{Ben
  Chamberlain}, \bibinfo{person}{Fabrizio Frasca}, \bibinfo{person}{Davide
  Eynard}, \bibinfo{person}{Federico Monti}, {and} \bibinfo{person}{Michael
  Bronstein}.} \bibinfo{year}{2020}\natexlab{}.
\newblock \showarticletitle{Temporal graph networks for deep learning on
  dynamic graphs}.
\newblock \bibinfo{journal}{\emph{arXiv preprint arXiv:2006.10637}}
  (\bibinfo{year}{2020}).
\newblock


\bibitem[\protect\citeauthoryear{Shamsi, Victor, Kantarcioglu, Gel, and
  Akcora}{Shamsi et~al\mbox{.}}{2022}]%
        {shamsi2022chartalist}
\bibfield{author}{\bibinfo{person}{Kiarash Shamsi}, \bibinfo{person}{Friedhelm
  Victor}, \bibinfo{person}{Murat Kantarcioglu}, \bibinfo{person}{Yulia Gel},
  {and} \bibinfo{person}{Cuneyt~G Akcora}.} \bibinfo{year}{2022}\natexlab{}.
\newblock \showarticletitle{Chartalist: Labeled graph datasets for utxo and
  account-based blockchains}.
\newblock \bibinfo{journal}{\emph{Advances in Neural Information Processing
  Systems}}  \bibinfo{volume}{35} (\bibinfo{year}{2022}),
  \bibinfo{pages}{34926--34939}.
\newblock


\bibitem[\protect\citeauthoryear{Sharma, Jiang, Bommannavar, Larson, and
  Lin}{Sharma et~al\mbox{.}}{2016}]%
        {sharma2016graphjet}
\bibfield{author}{\bibinfo{person}{Aneesh Sharma}, \bibinfo{person}{Jerry
  Jiang}, \bibinfo{person}{Praveen Bommannavar}, \bibinfo{person}{Brian
  Larson}, {and} \bibinfo{person}{Jimmy Lin}.} \bibinfo{year}{2016}\natexlab{}.
\newblock \showarticletitle{GraphJet: Real-time content recommendations at
  Twitter}.
\newblock \bibinfo{journal}{\emph{Proceedings of the VLDB Endowment}}
  \bibinfo{volume}{9}, \bibinfo{number}{13} (\bibinfo{year}{2016}),
  \bibinfo{pages}{1281--1292}.
\newblock


\bibitem[\protect\citeauthoryear{Shchur, Mumme, Bojchevski, and
  G{\"u}nnemann}{Shchur et~al\mbox{.}}{2018}]%
        {shchur2018pitfalls}
\bibfield{author}{\bibinfo{person}{Oleksandr Shchur},
  \bibinfo{person}{Maximilian Mumme}, \bibinfo{person}{Aleksandar Bojchevski},
  {and} \bibinfo{person}{Stephan G{\"u}nnemann}.}
  \bibinfo{year}{2018}\natexlab{}.
\newblock \showarticletitle{Pitfalls of graph neural network evaluation}.
\newblock \bibinfo{journal}{\emph{arXiv preprint arXiv:1811.05868}}
  (\bibinfo{year}{2018}).
\newblock


\bibitem[\protect\citeauthoryear{Tang, Qu, Wang, Zhang, Yan, and Mei}{Tang
  et~al\mbox{.}}{2015}]%
        {tang2015line}
\bibfield{author}{\bibinfo{person}{Jian Tang}, \bibinfo{person}{Meng Qu},
  \bibinfo{person}{Mingzhe Wang}, \bibinfo{person}{Ming Zhang},
  \bibinfo{person}{Jun Yan}, {and} \bibinfo{person}{Qiaozhu Mei}.}
  \bibinfo{year}{2015}\natexlab{}.
\newblock \showarticletitle{Line: Large-scale information network embedding}.
  In \bibinfo{booktitle}{\emph{Proceedings of the 24th international conference
  on world wide web}}. \bibinfo{pages}{1067--1077}.
\newblock


\bibitem[\protect\citeauthoryear{Trivedi, Farajtabar, Biswal, and Zha}{Trivedi
  et~al\mbox{.}}{2019}]%
        {trivedi2019dyrep}
\bibfield{author}{\bibinfo{person}{Rakshit Trivedi}, \bibinfo{person}{Mehrdad
  Farajtabar}, \bibinfo{person}{Prasenjeet Biswal}, {and}
  \bibinfo{person}{Hongyuan Zha}.} \bibinfo{year}{2019}\natexlab{}.
\newblock \showarticletitle{Dyrep: Learning representations over dynamic
  graphs}. In \bibinfo{booktitle}{\emph{International conference on learning
  representations}}.
\newblock


\bibitem[\protect\citeauthoryear{Vaswani}{Vaswani}{2017}]%
        {vaswani2017attention}
\bibfield{author}{\bibinfo{person}{A Vaswani}.}
  \bibinfo{year}{2017}\natexlab{}.
\newblock \showarticletitle{Attention is all you need}.
\newblock \bibinfo{journal}{\emph{Advances in Neural Information Processing
  Systems}} (\bibinfo{year}{2017}).
\newblock


\bibitem[\protect\citeauthoryear{Vaswani, Shazeer, Parmar, Uszkoreit, Jones,
  Gomez, Kaiser, and Polosukhin}{Vaswani et~al\mbox{.}}{2023}]%
        {transformer}
\bibfield{author}{\bibinfo{person}{Ashish Vaswani}, \bibinfo{person}{Noam
  Shazeer}, \bibinfo{person}{Niki Parmar}, \bibinfo{person}{Jakob Uszkoreit},
  \bibinfo{person}{Llion Jones}, \bibinfo{person}{Aidan~N. Gomez},
  \bibinfo{person}{Lukasz Kaiser}, {and} \bibinfo{person}{Illia Polosukhin}.}
  \bibinfo{year}{2023}\natexlab{}.
\newblock \bibinfo{title}{Attention Is All You Need}.
\newblock
\newblock
\showeprint[arxiv]{1706.03762}~[cs.CL]
\urldef\tempurl%
\url{https://arxiv.org/abs/1706.03762}
\showURL{%
\tempurl}


\bibitem[\protect\citeauthoryear{Wang, Chang, Li, Chu, Li, Zhang, He, Song,
  Zhou, and Yang}{Wang et~al\mbox{.}}{2021a}]%
        {wang2021tcl}
\bibfield{author}{\bibinfo{person}{Lu Wang}, \bibinfo{person}{Xiaofu Chang},
  \bibinfo{person}{Shuang Li}, \bibinfo{person}{Yunfei Chu},
  \bibinfo{person}{Hui Li}, \bibinfo{person}{Wei Zhang},
  \bibinfo{person}{Xiaofeng He}, \bibinfo{person}{Le Song},
  \bibinfo{person}{Jingren Zhou}, {and} \bibinfo{person}{Hongxia Yang}.}
  \bibinfo{year}{2021}\natexlab{a}.
\newblock \showarticletitle{Tcl: Transformer-based dynamic graph modelling via
  contrastive learning}.
\newblock \bibinfo{journal}{\emph{arXiv preprint arXiv:2105.07944}}
  (\bibinfo{year}{2021}).
\newblock


\bibitem[\protect\citeauthoryear{Wang, Xu, Wu, and Zhou}{Wang
  et~al\mbox{.}}{2014}]%
        {wang2014link}
\bibfield{author}{\bibinfo{person}{Peng Wang}, \bibinfo{person}{BaoWen Xu},
  \bibinfo{person}{YuRong Wu}, {and} \bibinfo{person}{XiaoYu Zhou}.}
  \bibinfo{year}{2014}\natexlab{}.
\newblock \showarticletitle{Link prediction in social networks: the
  state-of-the-art}.
\newblock \bibinfo{journal}{\emph{arXiv preprint arXiv:1411.5118}}
  (\bibinfo{year}{2014}).
\newblock


\bibitem[\protect\citeauthoryear{Wang, Lyu, Li, Xia, Yang, Wang, Wang, Cui,
  Yang, Sun, et~al\mbox{.}}{Wang et~al\mbox{.}}{2021b}]%
        {wang2021apan}
\bibfield{author}{\bibinfo{person}{Xuhong Wang}, \bibinfo{person}{Ding Lyu},
  \bibinfo{person}{Mengjian Li}, \bibinfo{person}{Yang Xia},
  \bibinfo{person}{Qi Yang}, \bibinfo{person}{Xinwen Wang},
  \bibinfo{person}{Xinguang Wang}, \bibinfo{person}{Ping Cui},
  \bibinfo{person}{Yupu Yang}, \bibinfo{person}{Bowen Sun}, {et~al\mbox{.}}}
  \bibinfo{year}{2021}\natexlab{b}.
\newblock \showarticletitle{Apan: Asynchronous propagation attention network
  for real-time temporal graph embedding}. In
  \bibinfo{booktitle}{\emph{Proceedings of the 2021 international conference on
  management of data}}. \bibinfo{pages}{2628--2638}.
\newblock


\bibitem[\protect\citeauthoryear{Wang, Chang, Liu, Leskovec, and Li}{Wang
  et~al\mbox{.}}{2022}]%
        {wang2022inductive}
\bibfield{author}{\bibinfo{person}{Yanbang Wang}, \bibinfo{person}{Yen-Yu
  Chang}, \bibinfo{person}{Yunyu Liu}, \bibinfo{person}{Jure Leskovec}, {and}
  \bibinfo{person}{Pan Li}.} \bibinfo{year}{2022}\natexlab{}.
\newblock \bibinfo{title}{Inductive Representation Learning in Temporal
  Networks via Causal Anonymous Walks}.
\newblock
\newblock
\showeprint[arxiv]{2101.05974}~[cs.LG]
\urldef\tempurl%
\url{https://arxiv.org/abs/2101.05974}
\showURL{%
\tempurl}


\bibitem[\protect\citeauthoryear{Wu, Pan, Chen, Long, Zhang, and Philip}{Wu
  et~al\mbox{.}}{2020}]%
        {wu2020comprehensive}
\bibfield{author}{\bibinfo{person}{Zonghan Wu}, \bibinfo{person}{Shirui Pan},
  \bibinfo{person}{Fengwen Chen}, \bibinfo{person}{Guodong Long},
  \bibinfo{person}{Chengqi Zhang}, {and} \bibinfo{person}{S~Yu Philip}.}
  \bibinfo{year}{2020}\natexlab{}.
\newblock \showarticletitle{A comprehensive survey on graph neural networks}.
\newblock \bibinfo{journal}{\emph{IEEE transactions on neural networks and
  learning systems}} \bibinfo{volume}{32}, \bibinfo{number}{1}
  (\bibinfo{year}{2020}), \bibinfo{pages}{4--24}.
\newblock


\bibitem[\protect\citeauthoryear{Xu, Ruan, Korpeoglu, Kumar, and Achan}{Xu
  et~al\mbox{.}}{2020}]%
        {xu2020inductive}
\bibfield{author}{\bibinfo{person}{Da Xu}, \bibinfo{person}{Chuanwei Ruan},
  \bibinfo{person}{Evren Korpeoglu}, \bibinfo{person}{Sushant Kumar}, {and}
  \bibinfo{person}{Kannan Achan}.} \bibinfo{year}{2020}\natexlab{}.
\newblock \showarticletitle{Inductive representation learning on temporal
  graphs}.
\newblock \bibinfo{journal}{\emph{arXiv preprint arXiv:2002.07962}}
  (\bibinfo{year}{2020}).
\newblock


\bibitem[\protect\citeauthoryear{Xu, Li, Tian, Sonobe, Kawarabayashi, and
  Jegelka}{Xu et~al\mbox{.}}{2018}]%
        {xu2018representation}
\bibfield{author}{\bibinfo{person}{Keyulu Xu}, \bibinfo{person}{Chengtao Li},
  \bibinfo{person}{Yonglong Tian}, \bibinfo{person}{Tomohiro Sonobe},
  \bibinfo{person}{Ken-ichi Kawarabayashi}, {and} \bibinfo{person}{Stefanie
  Jegelka}.} \bibinfo{year}{2018}\natexlab{}.
\newblock \showarticletitle{Representation learning on graphs with jumping
  knowledge networks}. In \bibinfo{booktitle}{\emph{International conference on
  machine learning}}. PMLR, \bibinfo{pages}{5453--5462}.
\newblock


\bibitem[\protect\citeauthoryear{Yang, Cohen, and Salakhudinov}{Yang
  et~al\mbox{.}}{2016}]%
        {yang2016revisiting}
\bibfield{author}{\bibinfo{person}{Zhilin Yang}, \bibinfo{person}{William
  Cohen}, {and} \bibinfo{person}{Ruslan Salakhudinov}.}
  \bibinfo{year}{2016}\natexlab{}.
\newblock \showarticletitle{Revisiting semi-supervised learning with graph
  embeddings}. In \bibinfo{booktitle}{\emph{International conference on machine
  learning}}. PMLR, \bibinfo{pages}{40--48}.
\newblock


\bibitem[\protect\citeauthoryear{Yi, Peng, Zheng, Mo, Wei, Ye, Zixuan, and
  Huang}{Yi et~al\mbox{.}}{2025}]%
        {yi2025tgbseq}
\bibfield{author}{\bibinfo{person}{Lu Yi}, \bibinfo{person}{Jie Peng},
  \bibinfo{person}{Yanping Zheng}, \bibinfo{person}{Fengran Mo},
  \bibinfo{person}{Zhewei Wei}, \bibinfo{person}{Yuhang Ye},
  \bibinfo{person}{Yue Zixuan}, {and} \bibinfo{person}{Zengfeng Huang}.}
  \bibinfo{year}{2025}\natexlab{}.
\newblock \showarticletitle{{TGB}-Seq Benchmark: Challenging Temporal {GNN}s
  with Complex Sequential Dynamics}. In \bibinfo{booktitle}{\emph{The
  Thirteenth International Conference on Learning Representations}}.
\newblock
\urldef\tempurl%
\url{https://openreview.net/forum?id=8e2LirwiJT}
\showURL{%
\tempurl}


\bibitem[\protect\citeauthoryear{Yu, Sun, Du, and Lv}{Yu et~al\mbox{.}}{2023}]%
        {yu2023towards}
\bibfield{author}{\bibinfo{person}{Le Yu}, \bibinfo{person}{Leilei Sun},
  \bibinfo{person}{Bowen Du}, {and} \bibinfo{person}{Weifeng Lv}.}
  \bibinfo{year}{2023}\natexlab{}.
\newblock \showarticletitle{Towards better dynamic graph learning: New
  architecture and unified library}.
\newblock \bibinfo{journal}{\emph{Advances in Neural Information Processing
  Systems}}  \bibinfo{volume}{36} (\bibinfo{year}{2023}),
  \bibinfo{pages}{67686--67700}.
\newblock


\bibitem[\protect\citeauthoryear{Yu, Gong, Tao, Shen, Zhang, Yu, Liu, Zhang,
  Li, Luo, et~al\mbox{.}}{Yu et~al\mbox{.}}{2024a}]%
        {yu2024lsmgraph}
\bibfield{author}{\bibinfo{person}{Song Yu}, \bibinfo{person}{Shufeng Gong},
  \bibinfo{person}{Qian Tao}, \bibinfo{person}{Sijie Shen},
  \bibinfo{person}{Yanfeng Zhang}, \bibinfo{person}{Wenyuan Yu},
  \bibinfo{person}{Pengxi Liu}, \bibinfo{person}{Zhixin Zhang},
  \bibinfo{person}{Hongfu Li}, \bibinfo{person}{Xiaojian Luo}, {et~al\mbox{.}}}
  \bibinfo{year}{2024}\natexlab{a}.
\newblock \showarticletitle{LSMGraph: A High-Performance Dynamic Graph Storage
  System with Multi-Level CSR}.
\newblock \bibinfo{journal}{\emph{Proceedings of the ACM on Management of
  Data}} \bibinfo{volume}{2}, \bibinfo{number}{6} (\bibinfo{year}{2024}),
  \bibinfo{pages}{1--28}.
\newblock


\bibitem[\protect\citeauthoryear{Yu, Liao, and Luo}{Yu et~al\mbox{.}}{2024b}]%
        {yu2024genti}
\bibfield{author}{\bibinfo{person}{Zihao Yu}, \bibinfo{person}{Ningyi Liao},
  {and} \bibinfo{person}{Siqiang Luo}.} \bibinfo{year}{2024}\natexlab{b}.
\newblock \showarticletitle{GENTI: GPU-powered Walk-based Subgraph Extraction
  for Scalable Representation Learning on Dynamic Graphs}.
\newblock \bibinfo{journal}{\emph{Proceedings of the VLDB Endowment}}
  \bibinfo{volume}{17}, \bibinfo{number}{9} (\bibinfo{year}{2024}),
  \bibinfo{pages}{2269--2278}.
\newblock


\bibitem[\protect\citeauthoryear{Zhang, Li, Zhang, Qin, Qin, and Wang}{Zhang
  et~al\mbox{.}}{2024}]%
        {zhang2024efficient}
\bibfield{author}{\bibinfo{person}{Yalong Zhang}, \bibinfo{person}{Rong-Hua
  Li}, \bibinfo{person}{Qi Zhang}, \bibinfo{person}{Hongchao Qin},
  \bibinfo{person}{Lu Qin}, {and} \bibinfo{person}{Guoren Wang}.}
  \bibinfo{year}{2024}\natexlab{}.
\newblock \showarticletitle{Efficient Algorithms for Pseudoarboricity
  Computation in Large Static and Dynamic Graphs}.
\newblock \bibinfo{journal}{\emph{Proceedings of the VLDB Endowment}}
  \bibinfo{volume}{17}, \bibinfo{number}{11} (\bibinfo{year}{2024}),
  \bibinfo{pages}{2722--2734}.
\newblock


\bibitem[\protect\citeauthoryear{Zheng, Wei, and Liu}{Zheng
  et~al\mbox{.}}{2023}]%
        {DBLP:journals/pvldb/ZhengWL23}
\bibfield{author}{\bibinfo{person}{Yanping Zheng}, \bibinfo{person}{Zhewei
  Wei}, {and} \bibinfo{person}{Jiajun Liu}.} \bibinfo{year}{2023}\natexlab{}.
\newblock \showarticletitle{Decoupled Graph Neural Networks for Large Dynamic
  Graphs}.
\newblock \bibinfo{journal}{\emph{Proc. {VLDB} Endow.}} \bibinfo{volume}{16},
  \bibinfo{number}{9} (\bibinfo{year}{2023}), \bibinfo{pages}{2239--2247}.
\newblock
\urldef\tempurl%
\url{https://doi.org/10.14778/3598581.3598595}
\showDOI{\tempurl}


\bibitem[\protect\citeauthoryear{Zheng, Yi, and Wei}{Zheng
  et~al\mbox{.}}{2025}]%
        {zheng2025survey}
\bibfield{author}{\bibinfo{person}{Yanping Zheng}, \bibinfo{person}{Lu Yi},
  {and} \bibinfo{person}{Zhewei Wei}.} \bibinfo{year}{2025}\natexlab{}.
\newblock \showarticletitle{A survey of dynamic graph neural networks}.
\newblock \bibinfo{journal}{\emph{Frontiers of Computer Science}}
  \bibinfo{volume}{19}, \bibinfo{number}{6} (\bibinfo{year}{2025}),
  \bibinfo{pages}{1--18}.
\newblock


\bibitem[\protect\citeauthoryear{Zhou, Zheng, Nisa, Ioannidis, Song, and
  Karypis}{Zhou et~al\mbox{.}}{2022}]%
        {zhou2022tgl}
\bibfield{author}{\bibinfo{person}{Hongkuan Zhou}, \bibinfo{person}{Da Zheng},
  \bibinfo{person}{Israt Nisa}, \bibinfo{person}{Vasileios Ioannidis},
  \bibinfo{person}{Xiang Song}, {and} \bibinfo{person}{George Karypis}.}
  \bibinfo{year}{2022}\natexlab{}.
\newblock \showarticletitle{Tgl: A general framework for temporal gnn training
  on billion-scale graphs}.
\newblock \bibinfo{journal}{\emph{arXiv preprint arXiv:2203.14883}}
  (\bibinfo{year}{2022}).
\newblock


\bibitem[\protect\citeauthoryear{Zuo, Liu, Lin, Guo, Hu, and Wu}{Zuo
  et~al\mbox{.}}{2018}]%
        {zuo2018embedding}
\bibfield{author}{\bibinfo{person}{Yuan Zuo}, \bibinfo{person}{Guannan Liu},
  \bibinfo{person}{Hao Lin}, \bibinfo{person}{Jia Guo},
  \bibinfo{person}{Xiaoqian Hu}, {and} \bibinfo{person}{Junjie Wu}.}
  \bibinfo{year}{2018}\natexlab{}.
\newblock \showarticletitle{Embedding temporal network via neighborhood
  formation}. In \bibinfo{booktitle}{\emph{Proceedings of the 24th ACM SIGKDD
  international conference on knowledge discovery \& data mining}}.
  \bibinfo{pages}{2857--2866}.
\newblock


\end{thebibliography}

\clearpage
\appendix
\section{Appendix}

\subsection{Node Classification Task}\label{appx:node}
 We extend our proposed EAGLE  to the node classification task. Specifically, for each node \(v\) at time $t$, 
 we first select the  top-\(k_r\) most recent neighbors $N_v(t, k_r)$ following Section~\ref{sssec:time_model} and top-\(k_s\) nodes $N_v(\pi_v(t), k_s)$ with the largest T-PPR scores in Equation~\eqref{eq:ppr} in Section~\ref{sssec:structure_model}.
 Then, we first average  features $\mathbf{x}_{v_i}(t)$ and edge feature $\mathbf{e}_{v,v_i}(\tau_i)$ from \( N_v(t, k_r) \) to compute time-aware representation \( \mathbf{h}^r_v(t) \)as follows:
 \begin{align}
 	\mathbf{h}^r_v(t) = \frac{1}{k_r}\sum_{(v_i,\mathbf{e}_{v,v_i}(\tau_i),\tau_i) \in N_v(t, k_r)} [\mathbf{x}_{v_i}(t), \mathbf{e}_{v,v_i}(\tau_i)], \label{eq:letter:time_repre} 
 \end{align}
 Similarly, we average the node feature $\mathbf{x}_{v_i}(t)$ with transition probability $\pi_v(t)[v_i]$ from \( N_v(\pi_v(t), k_s) \) to compute structure-aware \( \mathbf{h}^s_v(t) \) as follows:
 \begin{align}
 	\mathbf{h}^s_v(t) = \frac{1}{k_s}\sum_{(v_i, \pi_v(t)[v_i]) \in N_v(\pi_v(t), k_s)} \mathbf{x}_{v_i}(t)\cdot \pi_v(t)[v_i], \label{eq:letter:struct_repre}
 \end{align}
 where $\pi_v(t)[v_i]$ is the transition probability from node $v$ to node $v_i$.
 Then, we directly apply a 2-layer MLPs to the concatenation of $[\mathbf{h}^r_v(t),\mathbf{h}^s_v(t)]$ to predict node label distribution $	\mathbf{y}^*_v(t) \in [0,1]^{|\mathcal{Y}|}$ for node $v$ as follows:
 \begin{align}
 	\mathbf{y}^*_v(t) = \sigma\big(\mathbf{W}_2\  \texttt{ReLU}(\mathbf{W}_1[\mathbf{h}^r_v(t), \mathbf{h}^r_v(t)]+\mathbf{b}_1) + \mathbf{b}_2\big), \label{eq:nc_score}
 \end{align}
 where $\sigma(\cdot)$ is \textsf{Softmax} function for single-label classification and is  \textsf{Sigmoid} function for multi-label  classification.

 To evaluate EAGLE, 
 we use four widely used temporal node classification benchmarks from TGB~\cite{huang2024temporal}, i.e., Trade~\cite{huang2024temporal}, Genre~\cite{huang2024temporal}, Reddit~\cite{huang2024temporal}, and Token~\cite{huang2024temporal}. 
 At any given time \(t\), each node \(v\) is associated with a ground truth label vector \(\mathbf{y}_v(t) \in [0,1]^{|\mathcal{Y}|}\), which denotes the probability of the node belonging to each category \(y \in \mathcal{Y}\), where $\left\| \mathbf{y}_v(t)\right\|_1=1$. The data descriptions are as follows:

 \begin{itemize}[leftmargin=*]

  	\item \textbf{Trade~\cite{macdonald2015rethinking}}: An agricultural trade dataset where nodes represent nations and edges denote trade values. The goal is to predict future trade proportions between nations.
 	
 	\item \textbf{Genre~\cite{kumar2019predicting, hidasi2012fast, bertin2011million}}: A music dataset with nodes for users and genres, and edges for user interactions. It aims to rank the genres a user will engage with most next week.
 	
 	\item \textbf{Reddit~\cite{nadiri2022large}}: A user–subreddit interaction dataset where nodes represent users and subreddits, and edges correspond to user posts. 
 	It predicts which type of  subreddits each user will interact with most next week.
 	
 	\item \textbf{Token~\cite{shamsi2022chartalist}}: A cryptocurrency transaction dataset with nodes for users and tokens, and edges for transactions. It predicts how frequently a user will interact with tokens next week.
 \end{itemize}

 \begin{table}[h]
 	\centering
 	\caption{Statistics of node classification datasets. 
 		\textit{\#Nodes} and \textit{\#Edges} represent the number of nodes and interactions.}
 \begin{tabular}{lccc}
 	\toprule
 	Dataset & \#Nodes & \#Edges & \#Time Steps \\
 	\midrule
 	Trade~\cite{macdonald2015rethinking}  & 255     & 468,245     & 32 \\
 	Genre~\cite{kumar2019predicting, hidasi2012fast, bertin2011million}  & 1,505   & 17,858,395  & 133,758 \\
 	Reddit~\cite{nadiri2022large} & 11,766  & 27,174,118  & 21,889,537 \\
 	Token~\cite{shamsi2022chartalist}  & 61,756  & 72,936,998  & 2,036,524 \\
 	\bottomrule
 \end{tabular}
\end{table}

In the node classification experiments, we compare EAGLE against effective TGNNs, including 
TGAT~\cite{xu2020inductive}, TGN~\cite{rossi2020temporal}, GMixer~\cite{congwe2023},  Zebra~\cite{li2023zebra}, and DyGFormer~\cite{yu2023towards}.
Following the Temporal Graph Benchmark protocol~\cite{huang2024temporal,huang2024tgb2}, 
we evaluate performance using three metrics: Normalized Discounted Cumulative Gain (NDCG@10), Mean Reciprocal Rank (MRR), and Hit Ratio (HR@10), all of which consider the relative order of candidate types $y \in \mathcal{Y}$ for each node. For example, in the genre dataset, NDCG@10 assesses how closely the model's top-10 predicted music genres correspond to the ground-truth ranking~\cite{huang2024temporal,huang2024tgb2}.

As shown in Table~\ref{tab:appx:nc_effectiveness},
EAGLE consistently outperforms all baseline models across all metrics, demonstrating its superior ability to balance short-term temporal recency and long-term structural patterns. For example, on the Trade dataset, 
EAGLE achieves an NDCG@10 of 0.8424, significantly higher than DyGFormer's 0.3880. 
In contrast, baseline models such as DyGFormer~\cite{yu2023towards}, Zebra~\cite{li2023zebra}, TGAT~\cite{xu2020inductive}, and GMixer~\cite{congwe2023} often fail to complete training on larger datasets (marked as TLE), reflecting their inefficiency.
As shown in Table~\ref{tab:appx:nc_efficiency}, EAGLE demonstrates a significant advantage in both efficiency and scalability, achieving a 300× speedup in training and nearly 50× speedup in inference compared to the state-of-the-art DyGFormer~\cite{yu2023towards}.
EAGLE   can handle large datasets where other models fail, demonstrating its higher practicability.

\begin{table*}[t]
	\caption{Effectiveness evaluation on node classification. Time Limit Exceed (TLE) indicates that the training process cannot be completed within 48 hours.}
	\setlength\tabcolsep{2.6pt}
	\newcolumntype{C}[1]{>{\centering\arraybackslash}p{#1}}
	\begin{tabular}{C{2.05cm}|C{1.1cm}|C{1.1cm}|C{1.1cm}|C{1.1cm}|C{1.1cm}|C{1.1cm}|C{1.1cm}|C{1.1cm}|C{1.1cm}|C{1.1cm}|C{1.1cm}|C{1.1cm}}
		\toprule
		\textbf{Dataset}   & \multicolumn{3}{c|}{\textbf{Trade}} & \multicolumn{3}{c|}{\textbf{Genre}} & \multicolumn{3}{c|}{\textbf{Reddit}} & \multicolumn{3}{c}{\textbf{Token}} \\ \midrule
		\textbf{Metric}    & \resizebox{0.063\textwidth}{\height}{\textbf{\textit{NDCG@10}}} & \textbf{\textit{MRR}} & \textbf{\textit{HR@10}} & \resizebox{0.063\textwidth}{\height}{\textbf{\textit{NDCG@10}}} & \textbf{\textit{MRR}} & \textbf{\textit{HR@10}} & \resizebox{0.063\textwidth}{\height}{\textbf{\textit{NDCG@10}}} & \textbf{\textit{MRR}} & \textbf{\textit{HR@10}} & \resizebox{0.063\textwidth}{\height}{\textbf{\textit{NDCG@10}}} & \textbf{\textit{MRR}} & \textbf{\textit{HR@10}} \\ \midrule
		\textbf{TGAT}      & 0.3743 & 0.2639 & 0.3515 & \underline{0.3672} & \underline{0.2480} & \underline{0.2992} & TLE & TLE & TLE & TLE & TLE & TLE \\  
		\textbf{TGN}       & 0.3742 & 0.2639 & 0.3512 & 0.3670 & 0.2479 & 0.2987 & \underline{0.3150} & \underline{0.2169} & \underline{0.2275} & TLE & TLE & TLE \\  
		\textbf{GMixer}    & 0.3739 & 0.2638 & 0.3514 & 0.3633 & 0.2452 & 0.2944 & TLE & TLE & TLE & TLE & TLE & TLE \\  
		\textbf{Zebra}     & 0.3747 & 0.2641 & 0.3522 & 0.3670 & 0.2477 & 0.2981 & 0.3020 & 0.2084 & 0.2156 & TLE & TLE & TLE \\
		\textbf{DyGFormer} & \underline{0.3880} & \underline{0.2713} & \underline{0.3689} & TLE & TLE & TLE & TLE & TLE & TLE & TLE & TLE & TLE \\ \midrule 
		\textbf{EAGLE}     & \textbf{0.8424} & \textbf{0.7459} & \textbf{0.7069} & \textbf{0.5140} & \textbf{0.3483} & \textbf{0.3848} & \textbf{0.5525} & \textbf{0.3877} & \textbf{0.3923} & \textbf{0.5108} & \textbf{0.4322} & \textbf{0.4202} \\ \bottomrule
	\end{tabular}
	\label{tab:appx:nc_effectiveness}
\end{table*}

\begin{table*}[]
	\caption{Efficiency evaluation on node classification. T-train (s) and T-infer (s) denote training and inference times. M-train (MB) and M-infer (MB) indicate peak GPU memory usage during training and inference. TLE indicates that the training process cannot be completed within 48 hours.}
	\setlength\tabcolsep{2.35pt}
	\begin{tabular}{c|c|c|c|c|c|c|c|c|c|c|c|c|c|c|c|c}
		\toprule
		\textbf{Dataset} & \multicolumn{4}{c|}{\textbf{Trade}} & \multicolumn{4}{c|}{\textbf{Genre}} & \multicolumn{4}{c|}{\textbf{Reddit}} & \multicolumn{4}{c}{\textbf{Token}} \\ \midrule
		\multirow{2}{*}{\textbf{Metric}} & \multicolumn{2}{c|}{\textbf{\textit{Time(s)}}} & \multicolumn{2}{c|}{\textbf{\textit{Mem(MB)}}} & \multicolumn{2}{c|}{\textbf{\textit{Time(s)}}} & \multicolumn{2}{c|}{\textbf{\textit{Mem(MB)}}} & \multicolumn{2}{c|}{\textbf{\textit{Time(s)}}} & \multicolumn{2}{c|}{\textbf{\textit{Mem(MB)}}} & \multicolumn{2}{c|}{\textbf{\textit{Time(s)}}} & \multicolumn{2}{c}{\textbf{\textit{Mem(MB)}}} \\ \cline{2-17} 
		& \textit{Train} & \textit{Infer} & \textit{Train} & \textit{Infer} & \textit{Train} & \textit{Infer} & \textit{Train} & \textit{Infer} & \textit{Train} & \textit{Infer} & \textit{Train} & \textit{Infer} & \textit{Train} & \textit{Infer} & \textit{Train} & \textit{Infer} \\ \midrule
		\textbf{TGAT} & 1083.83 & 20.22 & 986 & 719 & 49440.46 & 2880.49 & 1054 & 1220 & TLE & TLE & TLE & TLE & TLE & TLE & TLE & TLE \\
		\textbf{TGN} & 286.82 & 2.88 & 628 & 498 & 8189.60 & 135.37 & 820 & 854 & 105691.64 & 760.15 & 2320 & 2108 & TLE & TLE & TLE & TLE \\
		\textbf{Gmixer} & 982.07 & 8.51 & 820 & 620 & 34916.64 & 606.49 & 876 & 902 & TLE & TLE & TLE & TLE & TLE & TLE & TLE & TLE \\
		\textbf{Zebra} &  {165.34} &  {1.42} &    {412} &  {345} &  {3205.79} &  {103.84} &  {756} &  {771} &  {46757.61} &  {557.32} &  {1982} &  {1734} & TLE & TLE & TLE & TLE \\
		\textbf{DyGFormer} & 6742.65 & 29.53 & 1168 & 730 & TLE & TLE & TLE & TLE & TLE & TLE & TLE & TLE & TLE & TLE & TLE & TLE \\ \midrule
		\textbf{EAGLE} & \textbf{15.65} & \textbf{0.62} & \textbf{330} & \textbf{219} & \textbf{265.76} & \textbf{28.65} & \textbf{470} & \textbf{485} & \textbf{2010.64} & \textbf{174.66} & \textbf{1468} & \textbf{1070} & \textbf{2634.85} & \textbf{233.45} & \textbf{3726} & \textbf{2420} \\ \bottomrule
	\end{tabular}
	\label{tab:appx:nc_efficiency}
\end{table*}

\end{document}